\pdfoutput=1  % arXiv: compile with pdflatex
\PassOptionsToPackage{table}{xcolor}  % supplement uses \rowcolor
\documentclass[conference]{IEEEtran}
\usepackage[T1]{fontenc}
\usepackage{amsmath,amssymb,booktabs,graphicx,xcolor,microtype,xspace, cite,tikz}
\usetikzlibrary{arrows.meta, positioning, calc}
\usepackage[hidelinks]{hyperref}
\usepackage{cleveref}
\crefname{figure}{Fig.}{Figs.}\Crefname{figure}{Fig.}{Figs.}
\crefname{table}{Table}{Tables}\Crefname{table}{Table}{Tables}
\crefname{section}{Section}{Sections}\Crefname{section}{Section}{Sections}
\graphicspath{{figs/}}

\usetikzlibrary{decorations.pathreplacing}
\usepackage[utf8]{inputenc}
\usepackage{enumitem}
\setlist[itemize]{leftmargin=*,itemsep=1pt,topsep=2pt,parsep=0pt}
\crefname{subsection}{Section}{Sections}\Crefname{subsection}{Section}{Sections}  % used by the supplement; the main paper does not \cref subsections
\newcommand{\gist}[1]{\noindent\textit{#1}\par\smallskip}   % one-line summary at the start of a section
\newcommand{\hl}{\rowcolor{black!8}}                        % highlights the row that the main paper uses
\makeatletter
\let\SuppSavedMaketitle\maketitle
\let\SuppSavedAtMaketitle\@maketitle
\makeatother
\newskip\SuppTextfloatsep \newskip\SuppDbltextfloatsep \newskip\SuppFloatsep

\begin{document}
\SuppTextfloatsep=\textfloatsep \SuppDbltextfloatsep=\dbltextfloatsep \SuppFloatsep=\floatsep  % class defaults, restored for the supplement
\setlength{\textfloatsep}{6pt plus 2pt minus 2pt}\setlength{\dbltextfloatsep}{8pt plus 2pt minus 4pt}\setlength{\floatsep}{6pt plus 2pt minus 2pt}
% \title{One Fringe Image, Sub-Millimetre Depth: Multi-Carrier Phase Matching on Top of a Depth Network}
% \title{MCPM-SS-PSP: Multi-Carrier Phase Mathcing -- Single Shot -- Phase Shift Profilometry}
\title{PG-MCPM: Physics-Guided Multi-Carrier Phase Matching with Learned Depth for Single-Shot 3D Profilometry}

\author{\IEEEauthorblockN{Rhys G. Evans\textsuperscript{a,*}, Siddhartha Mishra\textsuperscript{b}, and Sam Van der Jeught\textsuperscript{a}}
\IEEEauthorblockA{\textsuperscript{a}Industrial Vision Lab (InViLab), Faculty of Applied Engineering, University of Antwerp,\\
Groenenborgerlaan 179, Antwerp, 2020, Belgium\\
\textsuperscript{b}Seminar for Applied Mathematics (SAM), Department of Mathematics, ETH Zurich,\\
R\"amistrasse 101, Zurich, 8092, Switzerland\\
\textsuperscript{*}Corresponding author: R.\,G.~Evans, \href{mailto:rhys.evans@uantwerpen.be}{rhys.evans@uantwerpen.be}}}

\maketitle

\begin{abstract}
Single-shot fringe projection profilometry (SS-FPP) must recover depth from a single image, despite fringe ambiguity and the limited accuracy of depth networks. These limitations are addressed by combining multi-frequency, multi-orientation fringe encoding with a fine-tuned Depth Anything V2 network. The resulting image contains four superimposed fringes (two orientations and two periods), from which the network predicts a coarse metric depth map and the wrapped phase of each carrier. Subsequently, ray-wise phase matching identifies, for each pixel, the depth closest to the coarse estimate that best explains all four phases through the calibrated camera--projector geometry. A neighbour-consistency test identifies reliably unwrapped pixels (the trusted mask), after which the remaining pixels are re-matched.\\
The method is evaluated on illuminated pixels with available reference depth, at a working distance of about 0.35\,m. On two unseen physical objects, 86.7\% of pixels lie within 1\,mm of the reference (0.3\% of the working distance; 6.8\% using coarse depth alone), while the mean absolute error (MAE) decreases from 9.3 to 2.5\,mm (2.7\% to 0.7\% of the working distance). Of the reconstructed pixels, 91.2\% are trusted, with 92.2\% within 1\,mm (median error: 0.03\,mm), although 7.4\% exceed 5\,mm. New views of training objects achieve 95.8\% accuracy within 1\,mm. The matcher is tuned on four objects without retraining the network. A U-Net trained from scratch achieves comparable accuracy on the unseen objects (89.5\%), whereas the pre-trained model performs better on three of the four held-out objects.
\end{abstract}

\section{Introduction}\label{PG-MCPM-sec:introduction}
\noindent Fringe projection profilometry reconstructs depth from the deformation of a projected pattern \cite{gorthi2010fringe,zhang2018highspeed}. Conventional multi-shot methods project several phase-shifted patterns per frame; single-shot methods use one image, which suits moving objects and simple hardware but removes the redundancy that makes phase shifting accurate \cite{takeda1983fourier,vanderjeught2019deep,qian2020single,nguyen2020single,feng2019fringe,zuo2022deep}. Networks that regress depth directly from one fringe image reach centimetre-level accuracy or better on their training distribution, but tend to rely on object shape and context rather than on the fringe measurement \cite{evans2024deep,haroon2026diagnosing}. Networks that output the wrapped phase keep the measurement physics, but a single wrapped phase is periodic, so the fringe order is unknown and must come from additional information \cite{feng2019fringe,haroon2026repairing}.\\
This paper treats the two outputs as complementary and combines them geometrically. The depth prediction is accurate to about 10\,mm but coarse; the wrapped phases are precise to a fraction of a fringe but ambiguous. Their combination gives depth within a millimetre on most pixels, provided that the ambiguity period (66\,mm in depth at 0.35\,m, against 15\,mm for the two 16-pixel carriers alone) exceeds the error of the coarse depth: the coarse depth then selects the fringe order, and the phases place the depth within it. The contributions are:
\textbf{(1)} a single-shot pipeline in which one fine-tuned network (DA-V2 \cite{yang2024depthanythingv2}) predicts coarse depth and the wrapped phases of four superimposed carriers, and a ray-wise matcher fuses them without unwrapping logic;
\textbf{(2)} a neighbour-consistency test that marks reliable pixels without a reference and re-matches the others;
\textbf{(3)} matcher settings (prior width, phase noise) selected from four labelled physical objects and evaluated, without retraining, on objects not used for training or tuning;
\textbf{(4)} an oracle-input analysis of what limits the result: the wrapped-phase error, not the coarse depth; and \textbf{(5)} a comparison with a scratch-trained network, less robust when objects are left out.
The method is device-specific: the network and the matcher are tuned to one scanner, and generalisation is sought across objects measured with it. A new scanner needs its calibration, rendered training data and labelled scans of a few objects.

\section{Related Work}\label{PG-MCPM-sec:related-work}
\noindent \textbf{Single-shot fringe profilometry with networks.} Networks have been trained to regress depth from one fringe image \cite{vanderjeught2019deep,nguyen2020single,cai2024ttfdnet}, to predict the intermediate phase quantities of phase-shifting analysis \cite{feng2019fringe,qian2020single}, or to use the projected pattern design as a lever on accuracy \cite{evans2025pattern}. Studies with simulated sensors report that depth-regression networks fall back on shape priors in place of the fringe signal, and that predicting a phase quantity first reduces this \cite{haroon2026virtus,haroon2026diagnosing,haroon2026repairing}. The present pipeline takes the second route without image-domain unwrapping. Geometry-constrained unwrapping \cite{an2016pixelwise} resolves the fringe order from the calibrated geometry; the matcher does so for superimposed carriers with a learned depth prior.\\
\noindent \textbf{Depth foundation models.} Depth Anything and its successor \cite{yang2024depthanything,yang2024depthanythingv2} build on DINOv2 features \cite{oquab2024dinov2} with a dense-prediction decoder \cite{ranftl2021dpt}. A U-Net and a ViT without pre-training are compared with it (Sec.~\ref{PG-MCPM-subsec:combination-vs-parts}). Depth completion \cite{mmdcnet} needs sparse measured depth and is not comparable.

\section{Method}\label{PG-MCPM-sec:method}
\begin{figure*}[t]
\centering
\resizebox{0.8\textwidth}{!}{%
\begin{tikzpicture}[font=\scriptsize,
 L/.style={draw=blue!70!black,fill=blue!8,rounded corners=1pt,align=center,minimum height=8mm,minimum width=14mm,inner sep=1.5pt},
 N/.style={draw=blue!70!black,fill=blue!8,dashed,rounded corners=1pt,align=center,minimum height=8mm,minimum width=14mm,inner sep=1.5pt},
 A/.style={draw=black!55,dashed,rounded corners=1pt,align=center,minimum height=8mm,minimum width=14mm,inner sep=1.5pt},
 M/.style={draw=black!70,fill=black!8,rounded corners=1pt,align=center,minimum height=8mm,minimum width=14mm,inner sep=1.5pt},
 O/.style={draw=teal!60!black,fill=teal!10,rounded corners=1pt,align=center,minimum height=8mm,minimum width=14mm,inner sep=1.5pt},
 arr/.style={-{Stealth[length=1.5mm]},draw=black!70},node distance=4mm]
% main row: boxes are chained with a fixed gap, so arrows keep a visible length whatever the text width
\node[O] (in) at (0,0) {fringe image\\$1{\times}832{\times}1024$\\4 carriers};
\node[L,right=of in] (pe) {patch embed\\14\,px, 1 ch};
\node[L,right=of pe] (enc) {DINOv2 ViT-B\\12 blocks};
\node[L,right=of enc] (neck) {DPT neck\\4 scales};
\node[L,right=of neck] (c1) {head\\conv1};
\node[L,right=of c1] (c2) {head\\conv2\\(32 ch)};
\node[L,right=of c2] (c3) {conv3\\sigmoid\\$\times1.5$\,m};
\node[O,right=of c3] (z0) {coarse\\depth $z_c$};
\draw[arr] (in)--(pe); \draw[arr] (pe)--(enc); \draw[arr] (enc)--(neck); \draw[arr] (neck)--(c1); \draw[arr] (c1)--(c2); \draw[arr] (c2)--(c3); \draw[arr] (c3)--(z0);
% full-resolution fringe CNN feeding conv1 and conv2
\node[N] (f1) at ($(c1)!0.5!(c2)+(0,1.5)$) {fringe CNN, full res.\\140k, zero-init.};
\draw[arr] (in.north) |- (f1.west);
\draw[arr] (c1.north |- f1.south)--(c1.north);
\draw[arr] (c2.north |- f1.south)--(c2.north);
% phase path
\node[N] (dc) at ($(enc)+(0,-1.8)$) {U-Net, 5 levels\\base width 16, 1.9M};
\draw[arr] (in.south) |- (dc.west);
\node[N] (ph) at ($(c2)+(0,-1.8)$) {phase head\\conv $3{\times}3$, $1{\times}1$};
\draw[arr] (dc.east)--(ph.west);
\draw[arr] (c2.south)--(ph.north);
\node[O,minimum width=22mm] (phs) at ($(z0)+(0,-1.8)$) {8 ch: $\sin,\cos$\\of $x_{10.67},x_{16},$\\$y_{10.67},y_{16}$};
\draw[arr] (ph)--(phs);
% auxiliary order head (training only)
\node[A] (oh) at ($(c2)+(0,-3.3)$) {order head\\(auxiliary)\\2 conv $3{\times}3$, $1{\times}1$};
\draw[arr] ($(c2.south west)+(0.2,0)$) -- ++(0,-0.4) -- ++(-1.45,0) |- (oh.west);
\node[A] (uv) at ($(z0)+(0,-3.3)$) {$(u,v)/100$\\training only};
\draw[arr] (oh)--(uv);
% matcher, consistency step, output
\node[M] (mt) at ($(z0)+(2.5,-1.5)$) {ray-wise\\matching\\(Sec.~\ref{PG-MCPM-subsec:ray-wise-matching})};
\node[O,right=of mt] (cf) {consistency\\+ fill\\(Sec.~\ref{PG-MCPM-subsec:consistency-fill})};
\node[O,right=of cf] (out) {sub-mm depth\\+ trusted mask};
\node[A] (cal) at ($(mt)+(0,-1.8)$) {calibrated rig\\(camera, projector)};
\draw[arr] (z0.east)-|(mt.north);
\draw[arr] (phs.east)--(mt.west |- phs.east);
\draw[arr] (cal.north)--(mt.south);
\draw[arr] (mt)--(cf); \draw[arr] (cf)--(out);
\end{tikzpicture}}

\caption{\textbf{Pipeline from one fringe image to depth.} DA-V2 predicts the coarse depth $z_c$; a phase head, fed by the 32-channel decoder features and a small U-Net on the raw image, predicts the sine and cosine of the four wrapped phases; an order head shapes the features during training only. The ray-wise matcher combines both outputs with the calibrated geometry, and the consistency step marks trusted pixels and fills the rest. Solid blue: pretrained DA-V2 part, fine-tuned; dashed blue: new module; dashed grey: auxiliary or external.}
\label{PG-MCPM-fig:pipeline}
\end{figure*}
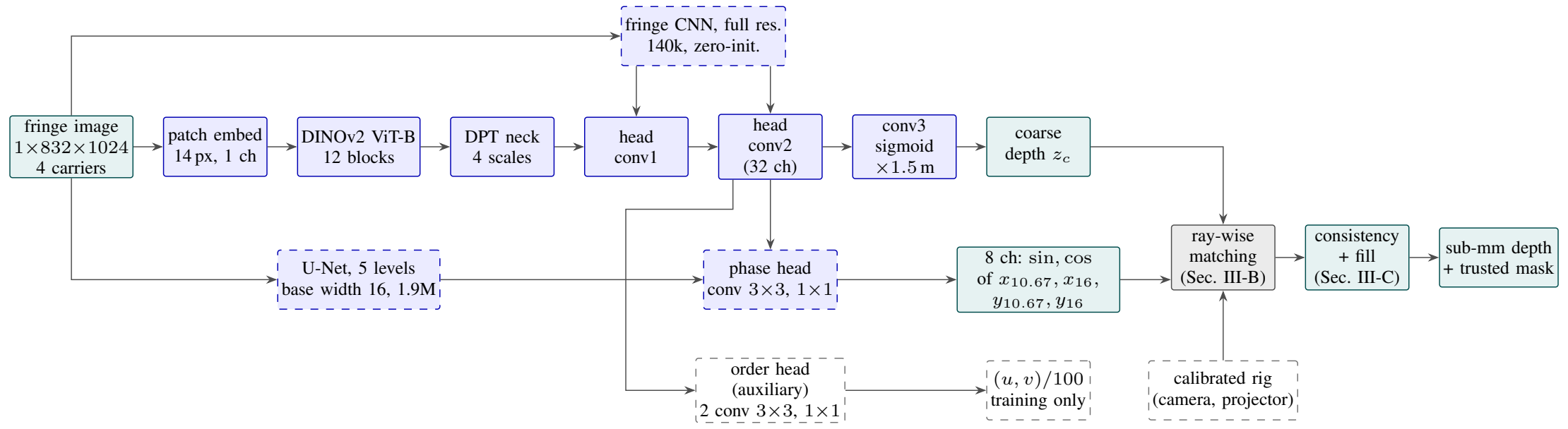

\noindent\Cref{PG-MCPM-fig:pipeline} gives the overview. The camera--projector rig is calibrated so that a camera ray with depth $z$ maps to a projector coordinate $(u(z),v(z))$. The pattern contains four carriers $c=1\ldots4$: sinusoids along the projector $x$ axis with periods 10.67 and 16 projector pixels and along the $y$ axis with the same periods. Their sum is the single projected image.

\subsection{Network}\label{PG-MCPM-subsec:network}
\noindent\textbf{Backbone.}
The backbone is DA-V2 Base \cite{yang2024depthanythingv2}, a DINOv2 ViT-B encoder \cite{oquab2024dinov2} with a DPT decoder \cite{ranftl2021dpt} and a metric-depth head ($d_{\max}=1.5$\,m), fine-tuned in full. A full-resolution CNN branch (140k parameters) feeds the first two decoder convolutions (\cref{PG-MCPM-fig:pipeline}). Its zero-initialised outputs leave the pretrained network's behaviour unchanged at the start of training.

\noindent\textbf{Heads.}
In addition to the depth head, a phase head predicts the sine and cosine of the four wrapped phases, from which the phase $\varphi_c=\operatorname{atan2}(s_c,c_c)$ and the modulation $m_c=\lVert(s_c,c_c)\rVert$ follow. It combines the 32-channel features of the second decoder convolution with those of a dilated-CNN or small five-level U-Net (feature-base width 16, 48 output channels, 1.9M parameters) applied directly to the input image. This path bypasses the encoder, whose 14\,px patches cannot resolve the 10.67\,px fringe period. An auxiliary order head predicts the absolute projector coordinates $(u,v)$ from the same features, the image, the pixel position and the detached coarse depth. It is used in training only.

\noindent\textbf{Training.}
Training proceeds in three stages: (1) train public DA-V2 weights on synthetic Blender data alone, 8 epochs; (2) train the whole pipeline on synthetic Blender data, 8 epochs; and (3) 200 epochs on real scans, with a masked L1 depth loss on the pixels that have a reference. The two training stages share one loss,
\begin{equation}
\mathcal L=\mathcal L_{\text{depth}}+\lambda_\varphi\mathcal L_{\text{phase}}+\lambda_o\mathcal L_{\text{order}},
\end{equation}
with $\lambda_\varphi=2$ and $\lambda_o=0.5$ in the real-data stage. All stages use AdamW, batch size 2 and a cosine schedule; the backbone learns at a tenth of the head rate ($5{\times}10^{-5}$, 8 epochs, synthetic stage; $2{\times}10^{-5}$, 200 epochs, real stage). The checkpoint is chosen on the illuminated validation MAE (the figurine's held-out positions).

\subsection{Ray-wise phase matching}\label{PG-MCPM-subsec:ray-wise-matching}
\noindent The network provides two cues: a coarse metric depth $z_c$ and four wrapped phases. The matcher searches each camera ray for the depth that agrees best with both.
For a pixel with camera ray $\mathbf r$, the expected phase of carrier $c$ at depth $z$ is
\begin{equation}
\hat\varphi_c(z)=2\pi\,\frac{q_c(z)-\tfrac12}{P_c},
\end{equation}
where $q_c(z)$ is the projector $x$ or $y$ coordinate (according to the carrier orientation) of the ray point at depth $z$, including projector lens distortion, and $P_c$ is the carrier period. The matcher minimises
\begin{equation}
J(z)=\frac{C}{\sigma_\varphi^{2}}\sum_{c}w_c\bigl[1-\cos(\varphi_c-\hat\varphi_c(z))\bigr]+\frac12\Bigl(\frac{z-z_c}{\sigma_c}\Bigr)^{2},
\end{equation}
with normalised weights $w_c$ and $C=4$. The first term is the wrapped-phase misfit of all carriers; the second is a Gaussian prior centred on the coarse depth with width $\sigma_c$. The search covers $z\in[z_c-R,\,z_c+R]$ with $R=\min(4\sigma_c,\,0.1\text{\,m})$, sampled every 0.5\,mm and then refined at 0.02\,mm around the minimum. Pixels with a phase magnitude below 0.25 in any carrier are excluded.\\
Wrapped phases alone cannot determine the fringe order, since all four carriers agree again at regular depth intervals and the phase term therefore has equally deep minima. At 0.35\,m these minima are 66\,mm apart with all four carriers, and 15\,mm with the two 16\,px carriers alone. The coarse depth must therefore be accurate to about half the four-carrier spacing ($\pm33$\,mm), so that the prior makes the minimum nearest to $z_c$ the deepest. The coarse depth thus selects the fringe order, and the phases then place the depth within that order to a fraction of a millimetre. Selecting the wrong minimum shifts the depth by one spacing. The margin between the best and second-best minimum of $J$ serves as a per-pixel confidence measure for the next step.

\subsection{Consistency and fill}\label{PG-MCPM-subsec:consistency-fill}
\noindent A matched depth is called \emph{trusted} if it lies within 1\,mm of the median of its $9{\times}9$ neighbourhood. Pixels that deviate by more than 3\,mm from that median are re-matched, with the median as the centre of a narrower prior (4\,mm, window $\pm20$\,mm), for up to three rounds. Pixels outside the matched set are filled with the coarse depth plus a correction field interpolated from trusted pixels by normalised convolution at three scales.

\subsection{Matcher settings}\label{PG-MCPM-subsec:matcher-settings}
\noindent The prior width $\sigma_c$ and the phase noise $\sigma_\varphi$ describe the scatter of the coarse depth and of the wrapped phase. The \emph{default} matcher (first version, fixed before any tuning) uses $\sigma_c=25$\,mm, $\sigma_\varphi=0.3$\,rad, and an edge dead zone that excludes pixels near depth edges from matching. The \emph{tuned} matcher, $\sigma_c=10$\,mm, $\sigma_\varphi=0.6$\,rad, no dead zone, was selected from a $3{\times}3{\times}2$ grid search ($\sigma_c\in\{10,15,25\}$\,mm, $\sigma_\varphi\in\{0.3,0.45,0.6\}$\,rad, dead zone on/off) on the four tuning objects (angel, restoration piece, bell, figurine). On unseen objects, it achieves 86.7\% vs 82.4\% for the default matcher; a 50\,mm prior gives 83.1\%. The prior width accounts for most of the gain (dilated-CNN model, prior 10\,mm, phase noise 0.3\,rad, dead zone on or off: 86.6--87.0\%). Priors of 5\,mm or less are narrower than the 9\,mm coarse-depth error, override the phases and fail (63--78\% at 0.6\,rad). Scores on the tuning objects (80.6––91.0\%) measure fit, not generalisation, and cannot rank the settings.

\section{Experimental Setup}\label{PG-MCPM-sec:experimental-setup}
\noindent\textbf{Scanner and data.} All data come from one camera--projector scanner. 15,000 synthetic images are rendered in Blender with the calibrated rig, using 3D-printed models \cite{Zhou2016_thingi10k} as objects. Real data are turntable scans of six objects (41 evaluated scans): angel, restoration piece, bell, figurine and fossil (8 each) and the benchmark (1).\\
\textbf{Scenes and objects.} A \emph{scene} is one scan (an object at one turntable position), an \emph{object} is the whole physical item. A test scan is an \emph{unseen scene} if its object occurs in training at other positions and an \emph{unseen object} if no scan of it is in training or in the matcher selection. Both are tested and are not pooled. \emph{Seen scenes:} the angel, restoration piece, bell, and figurine; scored seen scans are in training, so their numbers measure fit. \emph{Unseen scenes:} the seen figurine, evaluated on 8 held-out positions. \emph{Unseen objects:} the fossil and the benchmark.\\
\textbf{Scope and reference.} The target is depth in the illuminated region: only there does the image carry fringe information and a reference exist. It is 17\% of the evaluation grid (20\% on the unseen objects). The rest carries no phase information and is not scored. The \emph{evaluated pixels} are the illuminated pixels with a valid reference (87\% of the illuminated pixels on the unseen objects, 92\% on all scans). The reference depth is computed from the three-step phase of each carrier and the calibration. It shares its phase source with the method; a CAD model of the benchmark orbs checks it (reference minus CAD: MAE 0.20\,mm, 99.4\% within 1\,mm).\\
\textbf{Metrics.} All metrics are computed over the evaluated pixels. \emph{Within} denotes the share of pixels with an error below 1\,mm. MAE90 is the mean error over the 90\% of pixels with the smallest errors. It is reported because a few wrong-order pixels (errors above 5\,mm) can dominate the MAE. The trusted pixels (Sec.~\ref{PG-MCPM-subsec:consistency-fill}) are scored separately; the mask itself needs no reference.

\section{Results}\label{PG-MCPM-sec:results}
\subsection{The combination against each part alone}\label{PG-MCPM-subsec:combination-vs-parts}
\begin{figure*}[t]
\centering
\includegraphics[width=0.7\textwidth]{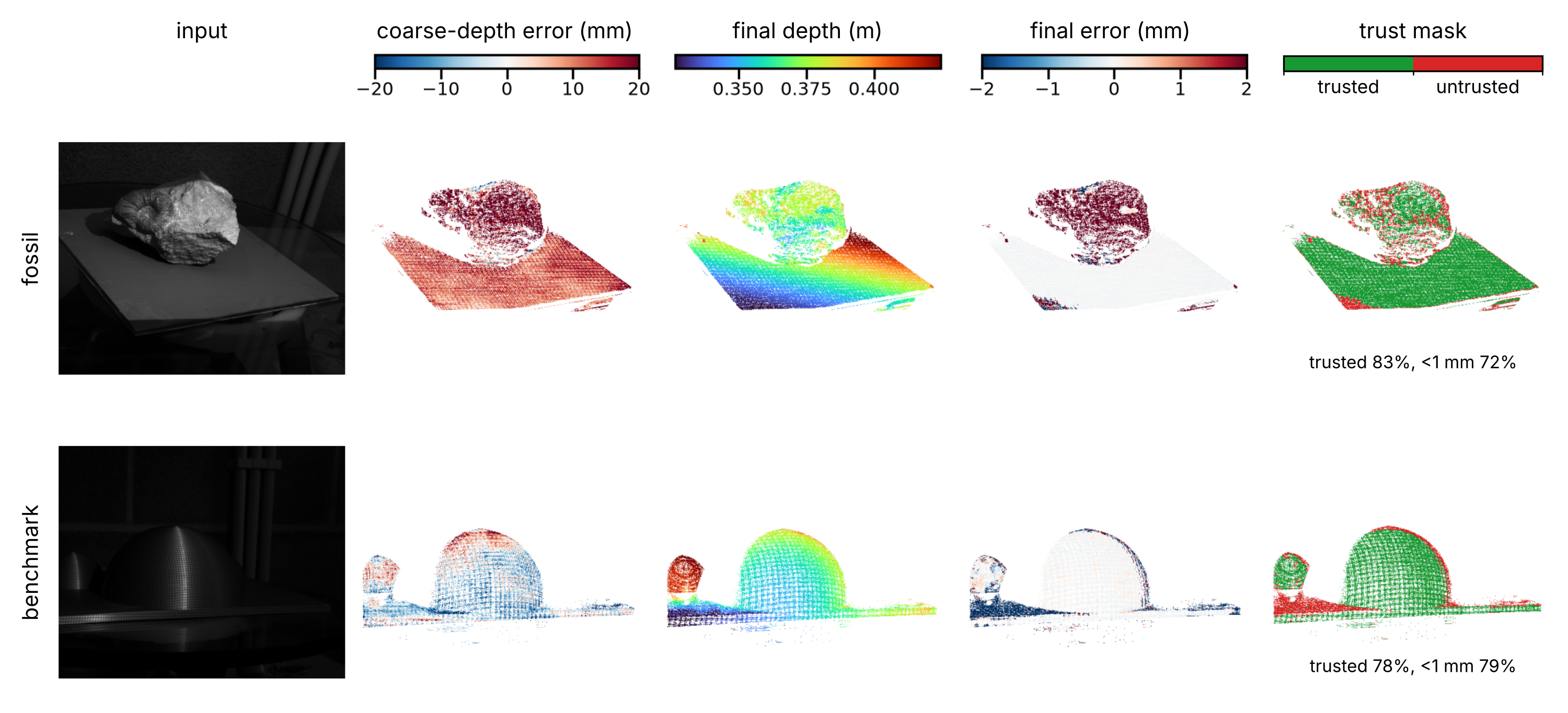}
\caption{Results on the first evaluation scan of each unseen object (fossil above, benchmark below), using the tuned matcher. Left to right: input, coarse-depth error, final depth, final error and trust mask; units and colour scales are given above each column. On the fossil, most remaining errors are wrong-order pixels on the body; this scan has 73\% within 1\,mm (87.2\% over all fossil scans).}
\label{PG-MCPM-fig:qualitative-results}
\end{figure*}

\noindent\Cref{PG-MCPM-tab:main-results,PG-MCPM-fig:qualitative-results} compare the outputs on the unseen objects. The coarse depth has a MAE of 9.3\,mm, with 6.8\% of the evaluated pixels within 1\,mm. A depth-only DA-V2 performs worse (12.1\,mm, 2.1\% trained on real data only; 10.5\,mm, 12.8\% with the synthetic stage), so the gain comes from the phases and the matcher rather than from a better coarse depth. With the matcher, FTP phases reach 58.2\% within 1\,mm and the U-Net network phases 82.4\% (default settings) and 86.7\%, with a MAE of 2.47\,mm and a MAE90 of 0.26\,mm. Of the evaluated pixels, 91.2\% are trusted. On these trusted pixels, 92.2\% are within 1\,mm (MAE 1.62\,mm, MAE90 0.04\,mm), but 7.4\% of them have errors above 5\,mm, and 76\% of those lie in coherent patches of at least 25 pixels, which the trust test cannot detect. On the figurine (an unseen scene, \Cref{PG-MCPM-tab:perset}) the matcher gives 0.58\,mm MAE and 95.8\% within 1\,mm, against 1.6\,mm and 45\% for a depth-only network.\\
\textbf{Network.} A U-Net of the same design trained from scratch with the same heads, data and recipe (\Cref{PG-MCPM-tab:main-results}) achieves a higher share within 1\,mm (89.5\% vs 86.7\%; 89.1--89.8\% over three seeds), a lower MAE90 (0.06 vs 0.26\,mm), and a larger trusted share (93.1\% vs 91.2\%). Its MAE is not lower (2.82\,mm, and 2.8--5.1\,mm over its seeds, against 2.47\,mm), because its failures are larger. \Cref{PG-MCPM-tab:perset} splits this by object: the U-Net is better on the fossil (90.9\% against 87.2\%) but worse on the single benchmark scan (72.8\% against 80.6\%; MAE 15.3\,mm, 15.3--46.9\,mm over its seeds, against 2.0\,mm). A ViT backbone without pre-training fails (9.3\% within 1\,mm, MAE 25.8\,mm on the unseen objects). Replacing the U-Net phase branch of the DA-V2 with a dilated CNN performs worse overall, see \Cref{PG-MCPM-tab:main-results}.\\
\textbf{Leave-one-object-out.} The lower block of \Cref{PG-MCPM-tab:perset} retrains both networks, using the dilated phase-branch variant for DA-V2. One of the four tuning objects is excluded from training and tested on both networks using its 8 scans. The matcher settings remain unchanged. With the DA-V2, the share within 1\,mm falls to 49--55\% on the angel, bell and restoration piece, and to 89.1\% on the figurine (94.7\% when it is in training, same variant). DA-V2 outperforms the scratch U-Net on three of the four held-out objects (angel, bell, figurine), with 3--5 times lower MAE; the scratch U-Net performs better on the restoration piece (69.1\% vs 54.3\%). The pre-trained backbone is thus not needed for typical pixels on unseen objects in \Cref{PG-MCPM-tab:main-results}, but it breaks down less on new objects.

\begin{table}[t]
\centering\small
\caption{Depth accuracy on evaluated pixels of the unseen objects (fossil and benchmark; 9 scans). MAE and MAE90 are reported in mm. FTP: Fourier-transform phases with the same matcher. ``U-Net branch'' denotes the phase branch of DA-V2; ``U-Net (scratch)'' denotes a separately trained network.}
\label{PG-MCPM-tab:main-results}
\setlength{\tabcolsep}{1.2pt}
\begin{tabular}{@{}lrrrr@{}}\toprule
Output & Within & MAE & MAE90 & Trusted\\\midrule
Depth-only, real data only & 2.1\% & 12.08 & -- & --\\
Depth-only, + synthetic stage & 12.8\% & 10.49 & -- & --\\
DA-V2 coarse depth & 6.8\% & 9.31 & -- & --\\
FTP phases + matcher & 58.2\% & 5.19 & 2.81 & --\\
U-Net branch, default matcher & 82.4\% & 5.07 & 1.10 & 87.5\%\\
\textbf{U-Net (scratch), same recipe} & \textbf{89.5\%} & \textbf{2.82} & \textbf{0.06} & \textbf{93.1\%}\\
\; trusted pixels only & 93.7\% & 1.76 & 0.04 & --\\
DA-V2, CNN branch (ours) & 85.9\% & 2.53 & 0.33 & 89.3\%\\
\textbf{DA-V2, U-Net branch (ours)} & \textbf{86.7\%} & \textbf{2.47} & \textbf{0.26} & \textbf{91.2\%}\\
\; trusted pixels only & 92.2\% & 1.62 & 0.04 & --\\\bottomrule
\end{tabular}
\end{table}

\begin{table}[t]
\centering\small
\caption{Accuracy within 1\,mm and MAE (mm) per test set for DA-V2 with the U-Net phase branch and a scratch U-Net. Upper block: all tuning objects included in training. Lower block: leave-one-object-out evaluation (DA-V2 with the dilated phase-branch variant), with each object tested on its 8 scans.}
\label{PG-MCPM-tab:perset}
\setlength{\tabcolsep}{3pt}
\begin{tabular}{@{}lrrrr@{}}\toprule
 & \multicolumn{2}{c}{DA-V2} & \multicolumn{2}{c}{U-Net (scratch)}\\
\cmidrule(lr){2-3}\cmidrule(lr){4-5}
Test set (scans) & Within & MAE & Within & MAE\\\midrule
Figurine (8) & 95.8\% & 0.58 & 96.7\% & 0.69\\
Fossil (8) & 87.2\% & 2.51 & 90.9\% & 1.80\\
Benchmark (1) & 80.6\% & 1.96 & 72.8\% & 15.29\\\midrule
\multicolumn{5}{@{}l}{\emph{Object left out of training}}\\
Angel (8) & 49.3\% & 8.51 & 42.3\% & 24.85\\
Bell (8) & 54.8\% & 9.14 & 30.2\% & 27.81\\
Restoration (8) & 54.3\% & 13.13 & 69.1\% & 9.69\\
Figurine (8) & 89.1\% & 2.18 & 67.4\% & 10.09\\\bottomrule
\end{tabular}
\end{table}

\subsection{Which parts of the combination matter}\label{PG-MCPM-subsec:ablation}
\noindent With the network outputs fixed, removing the consistency re-matching gives a smaller but more accurate trusted subset (86.4\% trusted, 94.4\% within 1\,mm) and fewer pixels within 1\,mm overall.\\
\textbf{All four carriers are needed.} Using the two 16-pixel carriers reduces the share within 1\,mm on unseen objects from 86.7\% to 58.9\%. Using three carriers loses 10--19\%, while the two 10.67-pixel carriers alone achieve 42.6\%.\\
\textbf{The carriers must be matched jointly.} Unwrapping each carrier independently against the coarse depth, as in classical reference-based phase unwrapping, yields 10--20\% within 1\,mm, because a coarse error of 7--10\,mm is too large to resolve a 10--15\,mm period; the four carriers together have a 66\,mm period that it can resolve (85.2\% within 1\,mm before clean-up).
\subsection{What limits the result}\label{PG-MCPM-subsec:error-sources}
\noindent Wrong-order pixels dominate the remaining errors: with the default matcher, 15.9\% of evaluated pixels across all scans exceed 5\,mm error, almost always by one fringe order. To isolate the source, each matcher input is separately replaced with reference data, while the other remains network-predicted. \emph{Ideal phases} are derived from reference depth and calibration. \emph{Reference coarse depth} adds 5\,mm Gaussian noise to the reference depth, within the $\pm33$\,mm tolerance. The network remains unchanged, isolating error sources rather than retraining potential.\\
\textbf{Ideal phases} (network coarse depth) remove about 80\% of the wrong-order pixels, reducing their share from 15.9\% to 2.9\%.\\
\textbf{Reference coarse depth} (network phases) lowers the share from 15.9\% to 14.4\%; with 15\,mm noise it rises to 18.0\%. Replacing both leaves 1.8\%. The network's coarse depth (MAE 5.7\,mm over all scans) is well within tolerance, so the phase error is the main source of wrong orders.\\
Wrong orders are frequent at low-modulation pixels (3.9\% of pixels, 60\% wrong) and depth edges (1.7\%, 37\%), but these hold about 15\% and 4\% of all wrong-order pixels; most occur at ordinary pixels with moderate phase error. A contrast-weighted phase loss and a refinement stage did not help the dilated-CNN variant. Network inference takes 0.15\,s and matching with consistency and fill 0.27\,s per image (RTX 4080), so the benefit of single-shot acquisition is the absence of motion between frames, not speed.

\section{Discussion and Conclusion}\label{PG-MCPM-sec:discussion}
\noindent We combined a fine-tuned depth network with ray-wise multi-carrier phase matching, so that a single fringe image gives both a depth map and a trust mask. The network supplies a coarse depth and the wrapped phases of four superimposed carriers. The coarse depth is too inaccurate on its own (MAE 9.3\,mm), but it is accurate enough to jointly select the fringe order of the four carriers. On the two unseen objects, 86.7\% of the evaluated pixels are within 1\,mm of the reference, 91.2\% are trusted, and 92.2\% of the trusted pixels are within 1\,mm. A U-Net trained from scratch reaches a slightly higher share on these objects (89.5\%), but its errors on a new object are larger, and it is worse than the pre-trained model on three of the four objects left out of training (\Cref{PG-MCPM-tab:perset}). The pre-trained model is therefore the more reliable choice for new objects.

{\small

}

%% =====================================================================
%% SUPPLEMENTARY MATERIAL (same PDF)
%% =====================================================================
\clearpage
% the stand-alone supplement was typeset without microtype
\microtypesetup{protrusion=false,expansion=false}
% float spacing as in the stand-alone supplement (class defaults)
\setlength{\textfloatsep}{\SuppTextfloatsep}\setlength{\dbltextfloatsep}{\SuppDbltextfloatsep}\setlength{\floatsep}{\SuppFloatsep}
% restart numbering, with S-prefixed numbers as in the stand-alone supplement
\setcounter{section}{0}\setcounter{subsection}{0}\setcounter{subsubsection}{0}
\setcounter{figure}{0}\setcounter{table}{0}\setcounter{equation}{0}\setcounter{footnote}{0}
\makeatletter
\providecommand{\thesectiondis}{}
\providecommand{\thesubsectiondis}{}
\renewcommand{\thesection}{S\arabic{section}}
\renewcommand{\thesectiondis}{\thesection.}
\renewcommand{\thesubsection}{\thesection.\arabic{subsection}}
\renewcommand{\thesubsectiondis}{\thesubsection}
\def\p@subsection{}
\makeatother
\renewcommand{\thetable}{S\arabic{table}}
\renewcommand{\thefigure}{S\arabic{figure}}
\renewcommand{\theequation}{S\arabic{equation}}
% unique hyperlink targets for the supplement
\renewcommand{\theHsection}{supp.\arabic{section}}
\renewcommand{\theHsubsection}{supp.\arabic{section}.\arabic{subsection}}
\renewcommand{\theHsubsubsection}{supp.\arabic{section}.\arabic{subsection}.\arabic{subsubsection}}
\renewcommand{\theHtable}{supp.\arabic{table}}
\renewcommand{\theHfigure}{supp.\arabic{figure}}
\renewcommand{\theHequation}{supp.\arabic{equation}}
\makeatletter
\let\maketitle\SuppSavedMaketitle
\let\@maketitle\SuppSavedAtMaketitle
\makeatother
\title{Supplementary Material for PG-MCPM: Physics-Guided Multi-Carrier Phase Matching with Learned Depth for Single-Shot 3D Profilometry}

\author{\IEEEauthorblockN{Rhys G. Evans\textsuperscript{a,*}, Siddhartha Mishra\textsuperscript{b}, and Sam Van der Jeught\textsuperscript{a}}
	\IEEEauthorblockA{\textsuperscript{a}Industrial Vision Lab (InViLab), Faculty of Applied Engineering, University of Antwerp,\\
		Groenenborgerlaan 179, Antwerp, 2020, Belgium\\
		\textsuperscript{b}Seminar for Applied Mathematics (SAM), Department of Mathematics, ETH Zurich,\\
		R\"amistrasse 101, Zurich, 8092, Switzerland\\
		\textsuperscript{*}Corresponding author: R.\,G.~Evans, \href{mailto:rhys.evans@uantwerpen.be}{rhys.evans@uantwerpen.be}}}

\maketitle

\noindent The main paper is self-contained. This document provides the evidence behind its statements: how the network is adapted and trained, how the matcher is set up, how it compares with other networks and classical methods, and how it behaves under noise and incorrect calibration. Each section starts with a one-line summary, and \cref{t:index} links the main paper's statements to the sections that support them.

\section{Reading guide}\label{s:guide}

\subsection{Conventions}
\noindent All tables follow the main paper's conventions.
\begin{itemize}
\item \textbf{Evaluated pixels.} The illuminated pixels with a valid reference depth. The reference is computed from the three-step phase of each carrier and the calibration (\cref{s:sets}).
\item \textbf{Metrics.} \emph{Within} is the share of evaluated pixels with an error below 1\,mm. MAE is in mm. MAE90 is the mean error over the 90\% of pixels with the smallest errors. \emph{Wrong order} is an error above 5\,mm. \emph{Trusted} pixels lie within 1\,mm of the median of their $9{\times}9$ neighbourhood; \emph{Trusted within} is Within among the trusted pixels.
\item \textbf{Scan sets.} \emph{All} is the 41 evaluated scans, \emph{Unseen} is the fossil and the benchmark (9 scans). \Cref{s:sets} defines seen scenes, unseen scenes and unseen objects.
\item \textbf{Matcher.} \emph{Tuned}: prior $\sigma_c=10$\,mm, phase noise $\sigma_\varphi=0.6$\,rad, no dead zone. \emph{Default}: 25\,mm, 0.3\,rad, dead zone on. Values are tuned unless the caption says ``default''.
\item \textbf{Shaded rows.} In tables that compare variants, a shaded row marks the setting or model that the main paper uses, so that the alternatives can be read against it.
\item \textbf{Intervals and seeds.} Intervals are 95\% bootstrap intervals over scans (2000 draws). Results are for one network seed unless a table shows several.
\end{itemize}

\subsection{Networks}
\noindent This document includes four networks. The first two share the DA-V2 Base backbone and the 140k-parameter full-resolution fringe branch of the main paper. They differ in the small convolutional path that feeds the phase head (the \emph{phase CNN}, defined in \cref{s:variant}).
\begin{itemize}
\item \textbf{DA-V2} is the final network of the main paper, with a five-level U-Net as phase CNN (1.9M parameters).
\item \textbf{Dilated} is the variant with a dilated stack as phase CNN (132k parameters). Most training ablations were run on it, so it is marked ``dilated'' wherever it is used. Both give similar results (\cref{t:arch}).
\item \textbf{U-Net (scratch)} (69.8M) and \textbf{ViT (scratch)} (40.6M) are the networks without pre-training of \cref{s:arch}.
\item The U-Net (124M) and ViT (89M) of \cref{s:syn} are earlier baselines for a different task (depth from one fringe) and are not the networks of \cref{s:arch}.
\end{itemize}

\subsection{Where to find what}
\noindent \Cref{t:index} lists where the main paper's statements are supported.
\begin{table}[!t]
\centering\small
\caption{Statements of the main paper and the sections that support them.}
\label{t:index}
\begin{tabular}{@{}p{0.64\columnwidth}p{0.31\columnwidth}@{}}\toprule
Statement in the main paper & Evidence\\\midrule
Unseen vs seen scenes and objects; pixel counts & \Cref{s:sets}\\
Tail of the trusted error; 76\% in coherent patches & \Cref{s:sets}, \cref{t:tail}\\
Adaptation of DA-V2 to fringe images & \Cref{s:syn}\\
Contribution of the training stages; depth-only baselines & \Cref{s:real}\\
Minima 66\,mm and 15\,mm apart; coarse depth within $\pm33$\,mm & \Cref{s:match}, \cref{f:cost}\\
Tuned vs default matcher; the 18-point grid & \Cref{s:settings}\\
Single carriers cannot unwrap; carrier subsets; carrier design & \Cref{s:single}, \cref{s:carrier}\\
Trust rule, clean-up, edge dead zone & \Cref{s:trust}\\
Seeds; leave-one-object-out; U-Net and ViT & \Cref{s:seeds,s:arch}\\
FTP phases; wrong-order pixels; oracle swaps & \Cref{s:ftp,s:phase}\\
Calibration, new rig, noise, CAD check, run time & \Cref{s:rob}\\\bottomrule
\end{tabular}
\end{table}

\section{Evaluation sets and error statistics}\label{s:sets}
\gist{Seen scenes measure fit; only the unseen objects and the leave-one-object-out runs measure generalisation to new objects. 91\% of the evaluated pixels on the unseen objects are trusted.}

\subsection{Scenes and objects}
\noindent A \emph{scene} is one scan: an object at one turntable position. An \emph{object} is the physical item. A test scan is an \emph{unseen scene} if its object occurs in training at other positions, and an \emph{unseen object} if no scan of it is in training or was used to select the matcher settings. Both kinds are tested and not pooled. The four \emph{tuning objects} are the angel, restoration piece, bell and figurine. \Cref{t:sets} lists the sets.
\begin{itemize}
\item \textbf{Seen scenes} (angel, restoration piece, bell): all scans are in training, so the numbers measure fit.
\item \textbf{Unseen scene} (figurine): trained on the other turntable positions, evaluated on eight held-out ones. It tests new views of a known object. These positions also serve as validation scans for checkpoint selection (\cref{s:real}), so the result is slightly optimistic.
\item \textbf{Unseen objects} (fossil, benchmark): the main result.
\item \textbf{Leave-one-object-out} (LOOO) removes a tuning object from training and turns it into an unseen object (\cref{s:seeds}).
\end{itemize}
Claims about new objects use only unseen objects and LOOO runs. The 41-scan row mixes all kinds.

\begin{table}[!t]
\centering\small
\caption{Real-scan evaluation sets (DA-V2, tuned matcher): share within 1\,mm with its 95\% interval over scans, MAE (mm) and trusted share. The benchmark is a single scan and has no interval.}
\label{t:sets}
\setlength{\tabcolsep}{2pt}\footnotesize
\begin{tabular}{@{}llrrrr@{}}\toprule
Category & Set (scans) & Within & [95\%] & MAE & Trusted\\\midrule
Seen scenes & Angel (8) & 85.5\% & [79.1, 90.9] & 1.28 & 82.9\%\\
 & Restoration (8) & 93.8\% & [92.5, 95.1] & 0.62 & 91.3\%\\
 & Bell (8) & 92.7\% & [90.8, 94.9] & 0.77 & 89.7\%\\
Unseen scene & Figurine, held-out (8) & 95.8\% & [95.1, 96.4] & 0.58 & 93.5\%\\
Unseen objects & Fossil (8) & 87.2\% & [80.1, 93.5] & 2.51 & 92.0\%\\
 & Benchmark (1) & 80.6\% & -- & 1.96 & 81.8\%\\
\hl & Fossil + benchmark (9) & 86.7\% & [79.8, 92.5] & 2.47 & 91.2\%\\\midrule
Mixed & All (41) & 91.7\% & [89.4, 93.5] & 1.15 & 90.9\%\\\bottomrule
\end{tabular}
\end{table}

\noindent The categories are not ordered by difficulty. The angel, a seen scene, scores below the unseen fossil because the objects differ in shape and surface. The figurine scores higher as an unseen scene (95.8\%) than as an unseen object (89.1\%, \cref{t:looo}). The reference shares its phase source with the method, so \cref{s:cad} checks it against a CAD model.

\subsection{Pixel accounting}
\noindent \Cref{t:counts} gives the pixel counts. The evaluated pixels are 92\% of the illuminated pixels on all scans and 87\% on the unseen objects; pixels without a reference are not scored. \Cref{t:acc} counts pixels within 1\,mm against three denominators: pixels with a reference, illuminated pixels, and all pixels in the evaluation grid. Of the 8.7 million grid pixels across the 41 scans, 14.0\% are final-depth pixels within 1\,mm, compared with 2.1\% for the coarse depth. For unseen objects, this is 15.0\% versus 1.2\%. The trusted pixels carry 96\% of the sub-millimetre pixels (1.17 of 1.22 million) and make up 90.9\% of the evaluated pixels.

\begin{table}[!t]
\centering\small
\caption{Pixel counts of the evaluation sets.}
\label{t:counts}
\setlength{\tabcolsep}{4pt}
\begin{tabular}{@{}lrrr@{}}\toprule
Scans & Grid & Illuminated & With reference\\\midrule
All (41) & 8,732,672 & 1,444,239 & 1,329,160\\
Unseen (9) & 1,916,928 & 378,932 & 330,605\\\bottomrule
\end{tabular}
\end{table}

\begin{table}[!t]
\centering\small
\caption{Pixels within 1\,mm (DA-V2, tuned matcher). Ref.\ = pixels with a reference; Illum.\ = illuminated pixels; Grid = all pixels of the evaluation grid (\cref{t:counts}). Counts in thousands.}
\label{t:acc}
\setlength{\tabcolsep}{2pt}
\begin{tabular}{@{}llrrrrr@{}}\toprule
Scans & Depth & Pixels & $<$1\,mm & of ref. & of illum. & of grid\\\midrule
All (41) & Coarse & 1329 & 181 & 13.6\% & 12.5\% & 2.1\%\\
 & Final & 1329 & 1219 & 91.7\% & 84.4\% & 14.0\%\\
 & Final, trusted & 1208 & 1166 & 87.7\% & 80.7\% & 13.3\%\\
\midrule
Unseen (9) & Coarse & 331 & 22 & 6.8\% & 5.9\% & 1.2\%\\
 & Final & 331 & 287 & 86.7\% & 75.6\% & 15.0\%\\
 & Final, trusted & 302 & 278 & 84.1\% & 73.4\% & 14.5\%\\\bottomrule
\end{tabular}
\end{table}

\noindent \Cref{t:err} gives the error statistics of the same pixel sets. Tail10 is the share of the summed error from the worst 10\% of pixels. For the final depth, it is 96\%, which is why the MAE (1.15\,mm) is about twenty times the MAE90 (0.06\,mm) and the two are reported together.

\begin{table}[!t]
\centering\small
\caption{Error statistics (mm) of the pixel sets of \cref{t:acc} (DA-V2, tuned matcher). Tail10 = share of the summed error from the worst 10\% of the pixels.}
\label{t:err}
\setlength{\tabcolsep}{1.5pt}\scriptsize
\begin{tabular}{@{}llrrrrrrr@{}}\toprule
Scans & Pixel set & MAE & RMSE & MAE90 & RMSE90 & Median & $<$1\,mm & Tail10\\\midrule
All & Coarse & 5.73 & 7.99 & 4.31 & 5.37 & 4.14 & 13.6\% & 32.2\%\\
 & Final & 1.15 & 4.84 & 0.06 & 0.09 & 0.04 & 91.7\% & 95.6\%\\
 & Trusted & 0.65 & 3.74 & 0.04 & 0.06 & 0.03 & 96.5\% & 94.2\%\\
 & Not trusted & 6.06 & 10.84 & 3.64 & 5.79 & 2.31 & 43.6\% & 46.0\%\\\midrule
Unseen & Coarse & 9.31 & 11.88 & 7.54 & 8.99 & 7.87 & 6.8\% & 27.2\%\\
 & Final & 2.47 & 7.77 & 0.26 & 1.20 & 0.03 & 86.7\% & 90.5\%\\
 & Trusted & 1.62 & 6.33 & 0.04 & 0.07 & 0.03 & 92.2\% & 97.5\%\\
 & Not trusted & 11.29 & 16.48 & 8.45 & 12.07 & 7.84 & 29.6\% & 32.7\%\\\bottomrule
\end{tabular}
\end{table}

\subsection{Tail of the trusted error}
\noindent \Cref{t:tail} gives percentiles of the absolute error and the spatial layout of the trusted pixels above 5\,mm. The trusted pixels are precise up to the 90th percentile (0.30\,mm on the unseen objects) and then have a tail of wrong-order pixels. Of the trusted pixels above 5\,mm, 76\% (unseen) and 73\% (all scans) lie in connected patches of at least 25 pixels, and 58\% and 57\% in patches of at least 100 pixels. Most of them are coherent regions that a neighbourhood test cannot reject.

\begin{table}[!t]
\centering\small
\caption{Absolute error percentiles (mm) of the evaluated and the trusted pixels (upper block), and the trusted pixels above 5\,mm, their share of the trusted pixels and the share lying in connected patches (8-neighbourhood) of at least 25 and 100 pixels (lower block). DA-V2, tuned matcher.}
\label{t:tail}
\setlength{\tabcolsep}{2pt}
\begin{tabular}{@{}llrrrrr@{}}\toprule
Scans & Pixels & p50 & p90 & p95 & p99 & Mean\\\midrule
Unseen & Evaluated & 0.03 & 9.50 & 22.06 & 34.97 & 2.47\\
 & Trusted & 0.03 & 0.30 & 14.94 & 33.20 & 1.62\\\midrule
All & Evaluated & 0.04 & 0.39 & 8.28 & 25.93 & 1.15\\
 & Trusted & 0.03 & 0.18 & 0.36 & 22.59 & 0.65\\\bottomrule
\end{tabular}
\vspace{2pt}

\begin{tabular}{@{}lrrrr@{}}\toprule
Scans & Pixels $>$5\,mm & Share & Patch $\ge25$ & Patch $\ge100$\\\midrule
Unseen & 22,388 & 7.42\% & 76.1\% & 58.2\%\\
All & 39,865 & 3.30\% & 72.8\% & 57.0\%\\\bottomrule
\end{tabular}
\end{table}

\section{Adapting DA-V2 to fringe depth regression}\label{s:syn}
\gist{The training recipe matters far more than the choice of adapter or fringe module. A depth-only network does not use the fringe as a measurement, which motivates the main paper's phase outputs.}

\noindent\textbf{Question.} Which adaptation choices matter when DA-V2 regresses depth from a single fringe image?

\begin{figure*}[t]
\centering\includegraphics[width=\textwidth]{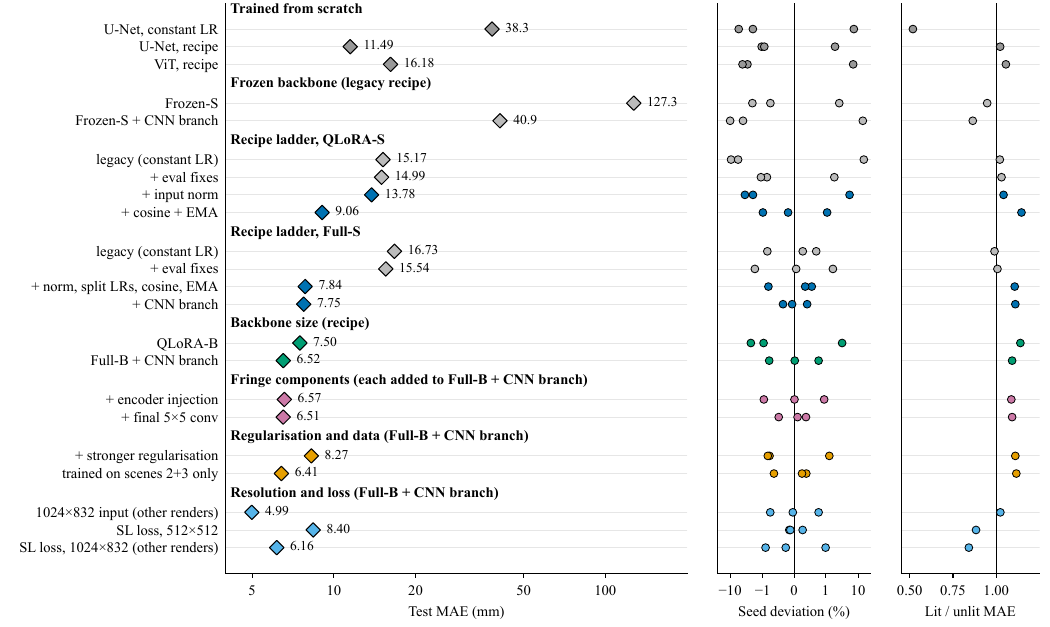}
\caption{Three-seed mean test MAE for the adaptation choices (left), per-seed deviation (middle) and the lit/unlit MAE ratio (right). Grey: legacy or from scratch; dark blue: recipe; green: backbone size; pink: fringe components; amber: regularisation and data; light blue: resolution and loss.}
\label{f:abl}
\end{figure*}

\noindent\textbf{Setting.}
\begin{itemize}
\item \textbf{Data.} 15,000 Blender renders of the calibrated rig (three scene groups, 80/10/10 split per image), one sinusoidal fringe of period 4--7 projector pixels, depth labels in a 0--1.5\,m window.
\item \textbf{Models.} DA-V2 Small (S) and Base (B) \cite{supp:yang2024depthanythingv2}, adapted frozen, with QLoRA (rank 32, 4-bit) \cite{supp:hu2022lora,supp:dettmers2023qlora}, or in full. The fringe-specific additions follow adapter designs for dense prediction \cite{supp:chen2023vitadapter}. Baselines are a U-Net \cite{supp:ronneberger2015unet} and a plain ViT \cite{supp:dosovitskiy2021vit}. Models are built with timm \cite{supp:wightman2019timm}, and the gradient-mixed loss follows \cite{supp:VanDerJeught2021_MixGE}.
\item \textbf{Metrics.} MAE in mm over all pixels, and the shape error left after a per-image scale and offset. Values are three-seed means with 95\% bootstrap intervals over test images.
\item \textbf{Caveat.} The split is per image, so this is a controlled benchmark and not a test of unseen objects. \Cref{t:syn} reports the full test set. The percentages in the text below are computed on the test images of scene groups 2 and 3, as in \cref{f:abl}, so they do not follow directly from \cref{t:syn}.
\end{itemize}

\begin{table}[!t]
\centering\small
\caption{Synthetic test MAE (mm), full test set, three-seed means. U-Net and ViT are the early baselines, not the networks of \cref{s:arch}. 1024\,px: $1024{\times}832$ input.}
\label{t:syn}
\setlength{\tabcolsep}{2pt}
\begin{tabular}{@{}lrrr@{}}\toprule
Model & Trainable & GPU-h & MAE\\\midrule
U-Net, from scratch & 124M & 20 & 9.91\\
ViT, from scratch & 89M & 12 & 13.78\\\midrule
DA-V2-S, QLoRA & 2.8M & 1.5 & 8.05\\
DA-V2-S, full + branch & 25M & 3.0 & 7.07\\
DA-V2-B, QLoRA & 5.6M & 2.6 & 6.89\\
DA-V2-B, full + branch & 97M & 4.8 & 6.10\\
DA-V2-B, full + branch, 1024 px & 97M & 20.5 & 5.03\\\bottomrule
\end{tabular}
\end{table}

\noindent\textbf{The training recipe dominates.} Compared with the legacy loop (constant rate, no schedule or EMA, last-epoch weights), the recipe reduces the test MAE of QLoRA-S by 40\% (15.0 to 9.1\,mm) and of full fine-tuning of S by 50\% (15.5 to 7.8\,mm). The recipe includes normalised input, a cosine schedule, EMA, separate rates for the head and backbone, and checkpoint selection on validation data. 96--97\% of the test images improve.

\noindent\textbf{Capacity and mode matter.} Base beats Small by 16\% (full) and 17\% (QLoRA). Base with QLoRA and 5.6M trainable parameters matches Small with full fine-tuning. A frozen backbone gives 127\,mm with the legacy recipe, and the CNN branch removes 68\% of that error. Both the U-Net and the ViT are clearly worse (\cref{t:syn}). The 512-pixel Base model runs in 13.5\,ms per frame (58\,ms at $1024{\times}832$).

\noindent\textbf{Fringe modules add little once the backbone is trained.} The full-resolution CNN branch changes the full-S error by $-1.2\%$ [$-2.0$, $-0.4$], encoder feature injection by $+0.9\%$ [$+0.1$, $+1.7$] and a final $5{\times}5$ refinement convolution by $-0.1\%$ [$-0.7$, $+0.5$]. All are below 2\%, against 40--50\% for the recipe. Stronger regularisation (drop-path 0.1, weight decay 0.05) raises the error by 27\%.

\noindent\textbf{A scanner-aware loss trades global for working-range error.} Weighting by depth range (0.30--0.60\,m) and illumination raises the plain MAE (7.41 against 6.10\,mm) and lowers it within the working range by 15\%.

\noindent\textbf{A depth network alone relies on shape.} A per-image scale and offset removes 46\% of the MAE (6.10 to 3.30\,mm), and lit pixels are not more accurate than unlit ones (lit/unlit ratio 1.0--1.15). A depth-only network does not use the fringe as a measurement. This motivates the phase outputs of the main paper.

\section{Training on real scans}\label{s:real}
\gist{The real-data stage is essential; the synthetic stages mainly improve the coarse depth, and the gain over a depth-only network comes from the phases and the matcher. Separate training of phase and depth does not beat joint training.}

\subsection{Phase head and phase CNN}\label{s:variant}
\begin{figure*}[!t]
\centering
\begin{tikzpicture}[font=\scriptsize,
 B/.style={draw=blue!60!black,fill=blue!6,rounded corners=1pt,minimum height=0.56cm,inner sep=2pt,align=center},
 W/.style={draw=black!70,fill=black!5,rounded corners=1pt,minimum height=0.56cm,inner sep=2pt,align=center},
 A/.style={-{Stealth[length=4pt]},line width=0.4pt},
 S/.style={-{Stealth[length=3pt]},line width=0.35pt,dashed,blue!60!black},
 T/.style={anchor=west,font=\footnotesize\bfseries}]
%% (a) phase head: three inputs, concatenated, three convolutions
\node[T] at (0,1.65) {(a) Phase head};
\node[W,minimum width=1.5cm] (img) at (0.8,-0.8) {image\\1 ch};
\node[B,minimum width=2.1cm] (cnn) at (3.6,0) {phase CNN, (b) or (c)\\48 ch};
\node[W,minimum width=2.1cm] (feat) at (3.6,0.8) {decoder (conv2) features\\32 ch};
\node[W,minimum width=2.1cm] (skip) at (3.6,-0.8) {image (skip)\\1 ch};
\node[W,minimum width=1.5cm,minimum height=2.3cm] (cat) at (6.2,0) {concat\\81 ch};
\node[B,minimum width=1.9cm] (c1) at (8.5,0) {conv $3{\times}3$\\64 ch, ReLU};
\node[B,minimum width=1.9cm] (c2) at (10.8,0) {conv $3{\times}3$\\64 ch, ReLU};
\node[B,minimum width=1.9cm] (c3) at (13.1,0) {conv $1{\times}1$\\8 ch};
\node[W,minimum width=1.9cm] (out) at (15.4,0) {$(\sin,\cos)$ of\\4 carriers};
\draw[A] (img.east)--(skip.west);
\fill (1.95,-0.8) circle (0.6pt);
\draw[A] (1.95,-0.8)|-(cnn.west);
\draw[A] (feat.east)--(feat.east -| cat.west);
\draw[A] (cnn.east)--(cnn.east -| cat.west);
\draw[A] (skip.east)--(skip.east -| cat.west);
\draw[A] (cat)--(c1); \draw[A] (c1)--(c2); \draw[A] (c2)--(c3); \draw[A] (c3)--(out);
%% (b) U-Net
\node[T] at (0,-1.9) {(b) U-Net phase CNN (final network), 1.94M parameters};
\foreach \i/\c/\x/\y in {1/16/1.6/-3.0, 2/32/3.4/-3.9, 3/64/5.2/-4.8, 4/128/7.0/-5.7, 5/256/8.8/-6.6}{
 \node[B,minimum width=0.8cm] (e\i) at (\x,\y) {\c};}
\foreach \i/\c/\x/\y in {4/128/10.6/-5.7, 3/64/12.4/-4.8, 2/32/14.2/-3.9, 1/16/16.0/-3.0}{
 \node[B,minimum width=0.8cm] (d\i) at (\x,\y) {\c};}
\draw[A] (e1)--(e2) node[midway,above right=-3pt] {pool}; \draw[A] (e2)--(e3); \draw[A] (e3)--(e4); \draw[A] (e4)--(e5);
\draw[A] (e5)--(d4) node[midway,above left=-3pt] {up}; \draw[A] (d4)--(d3); \draw[A] (d3)--(d2); \draw[A] (d2)--(d1);
\foreach \i in {1,2,3,4}{\draw[S] (e\i.east)--(d\i.west);}
\node[W] (uin) at (0.1,-3.0) {1 ch};
\node[W] (uout) at (17.5,-3.0) {48 ch};
\draw[A] (uin)--(e1); \draw[A] (d1)--(uout) node[midway,above] {$1{\times}1$};
\node[anchor=west,text width=6.2cm,align=left] at (0.2,-6.2) {Blue box: two $3{\times}3$ convolutions (no bias), each with batch norm and ReLU; the number is the output width. Dashed: skip connection (concatenation).};
%% (c) dilated stack
\node[T] at (0,-7.6) {(c) Dilated phase CNN (variant), 104k parameters};
\node[W,minimum width=1.2cm] (din) at (0.7,-8.8) {image\\1 ch};
\foreach \i/\d/\x in {1/1/3.0, 2/1/5.3, 3/2/7.6, 4/4/9.9, 5/8/12.2, 6/1/14.5}{
\node[B,minimum width=1.7cm] (l\i) at (\x,-8.8) {conv $3{\times}3$\\$d={\d}$};}
\node[W,minimum width=1.2cm] (dout) at (17.0,-8.8) {48 ch};
\draw[A] (din)--(l1); \foreach \i/\j in {1/2,2/3,3/4,4/5,5/6}{\draw[A] (l\i)--(l\j);} \draw[A] (l6)--(dout);
\draw[decorate,decoration={brace,amplitude=3pt,mirror},line width=0.4pt] (2.1,-9.3)--(15.4,-9.3) node[midway,below=4pt] {48 channels in every layer, ReLU, no normalisation, dilation $d$; receptive field 35\,px};
\end{tikzpicture}
\caption{Structure of the phase head and of the two phase CNNs. (a) The phase CNN output, the 32-channel features of the second decoder convolution and the image are concatenated and mapped to the eight phase outputs. (b) Five-level U-Net with base width 16, used in the final network. (c) Dilated stack of the dilated variant.}
\label{f:struct}
\end{figure*}
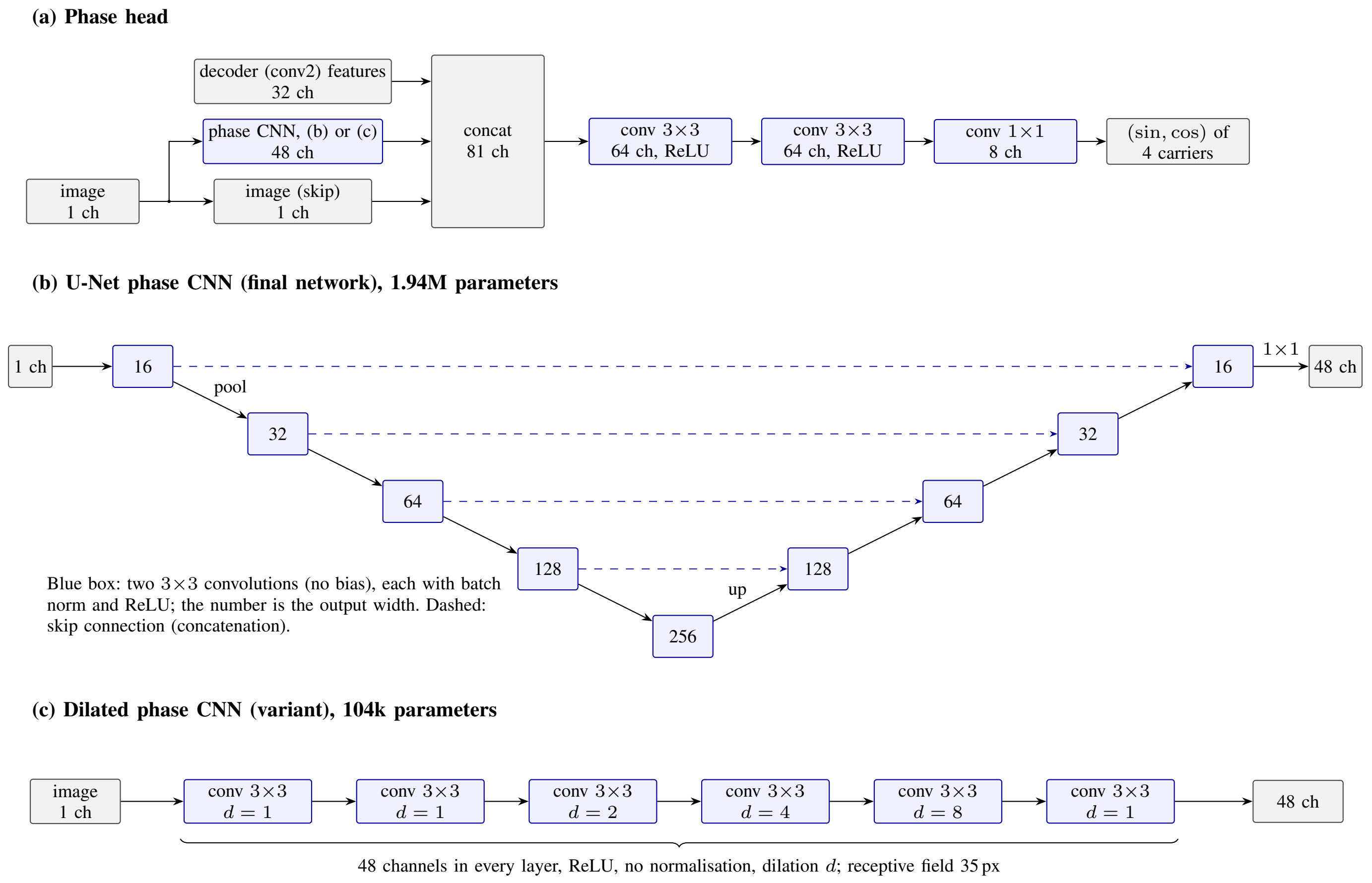

\noindent Two small convolutional paths operate on the raw image alongside the pre-trained network (\cref{f:struct}). The \emph{fringe branch} (135k parameters) is part of both networks. It consists of a two-convolution stem at full resolution (32 channels), a stride-2 stage of three convolutions (64 channels) and a fusion convolution (32 channels), each with $3{\times}3$ kernels, group normalisation and ReLU. Its features are added to the DA-V2 head through zero-initialised $1{\times}1$ projections, before \texttt{head.conv1} (64 channels, half resolution) and before \texttt{head.conv2} (32 channels, full resolution). At initialisation, the network therefore equals the plain DA-V2.

\noindent The \emph{phase CNN} reads the input image (one channel) and produces 48 channels at full resolution. These are concatenated with the 32-channel features of \texttt{head.conv2} and with the image itself (81 channels in total) and passed to the phase head: a $3{\times}3$ convolution with 64 channels and ReLU, a second one, and a $1{\times}1$ convolution to the eight outputs $(\sin\varphi_c,\cos\varphi_c)$, $c\in\{x_{10.67},x_{16},y_{10.67},y_{16}\}$. The two phase-CNN variants differ as follows.
\begin{itemize}
\item \textbf{U-Net (final network, 1.94M parameters).} Five levels with channel widths 16, 32, 64, 128 and 256. Each level is two $3{\times}3$ convolutions without bias, each followed by batch normalisation and ReLU. The encoder halves the resolution by $2{\times}2$ max-pooling. The decoder doubles it with a $2{\times}2$ transposed convolution that halves the channels, concatenates the skip connection of the same level and applies the double convolution. A final $1{\times}1$ convolution gives the 48 output channels, without activation. No pre-padding and no coordinate channels. This is the design of the scratch U-Net of \cref{s:arch}, with base width 16 instead of 64 and the heads removed.
\item \textbf{Dilated stack (dilated variant, 104k parameters).} Six $3{\times}3$ convolutions with ReLU and 48 channels each, at full resolution, with dilations 1, 1, 2, 4, 8 and 1, and without normalisation. The receptive field is 35\,px, a few fringe periods, so the local phase can be read from the image, which the encoder's 14\,px patches cannot resolve. With the wider input of the phase head it adds 132k parameters to the network.
\end{itemize}
The depth head, the order head, the data and the training are identical. The two give similar results (\cref{t:arch}): 86.7\% within 1\,mm on the unseen objects for the U-Net path and 85.9\% for the dilated stack. Most ablations of this section use the dilated variant, with the default matcher unless stated.

\subsection{Training stages}
\noindent The training chain has four stages, each starting from the previous stage's weights.
\begin{itemize}
\item \textbf{Stage 0.} The public Depth Anything V2 Base weights (DINOv2 ViT-B encoder with DPT decoder), which contain no fringe knowledge.
\item \textbf{Stage 1.} Synthetic renders of the calibrated rig with a crossed fringe pattern, 8 epochs. The from-scratch networks of \cref{s:arch} do not have this stage.
\item \textbf{Stage 2.} Synthetic renders with the four-carrier pattern, 8 epochs, learning rate $5{\times}10^{-5}$ for the head and a tenth of that for the backbone, phase and order weights 0.5, EMA 0.999. The network predicts depth, the four wrapped phases and the projector coordinates.
\item \textbf{Stage 3.} Real scans, 200 epochs, masked L1 depth loss on pixels with a reference, phase weight 2.0, order weight 0.5, learning rate $2{\times}10^{-5}$ (backbone factor 0.1). The checkpoint is selected on the illuminated MAE of the validation scans, which are the held-out positions of the figurine.
\end{itemize}
\Cref{t:stage} leaves out one stage at a time and always keeps the real stage: 0--2--3 (no crossed-fringe stage), 0--1--3 (no four-carrier stage) and 0--3 (real data only, from the public weights). ``Coarse MAE'' is the error of the network depth before the matching. A synthetic-only chain (0--1--2) evaluated zero-shot on the real scans has no pixel within 1\,mm and a coarse error of about 207\,mm, which shows the gap between the real scanner and the renders.

\begin{table}[!t]
\centering\small
\caption{Training stages (dilated, default matcher, all 41 scans).}
\label{t:stage}
\setlength{\tabcolsep}{2pt}
\begin{tabular}{@{}lrrrr@{}}\toprule
Stages & Trusted & Within & MAE & Coarse MAE\\\midrule
\hl Full chain, seed 0 & 82.7\% & 81.0\% & 5.29 & 6.57\\
Full chain, seed 1 & 82.9\% & 81.1\% & 5.25 & 6.43\\
Without crossed stage & 80.0\% & 78.9\% & 6.30 & 7.91\\
Without four-carrier stage & 77.1\% & 75.4\% & 6.13 & 7.33\\
Real data only (public DA-V2) & 68.5\% & 69.2\% & 10.61 & 13.19\\\bottomrule
\end{tabular}
\end{table}

\noindent The synthetic stages mainly improve the coarse depth. Real data alone gives a coarse MAE of 13.2\,mm, twice that of the full chain (6.6\,mm), and the lowest trusted share (68.5\% against 82.7\%). Of the two synthetic stages, the four-carrier stage matters more: leaving it out reduces the share within 1\,mm by 5.6\% (81.0\% to 75.4\%), against 2.1\% (to 78.9\%) for the crossed-fringe stage. The MAE does not separate them (6.13 and 6.30\,mm). The second seed of the full chain differs by 0.04\,mm in MAE and 0.1\% in Within, so these differences exceed seed variation. The crossed-fringe stage was trained for only 8 synthetic epochs, so its small contribution may partly reflect the short training.

\subsection{Depth-only baselines}
\noindent DA-V2 trained for depth only, with the same real data, epochs and recipe, is close to the reference on the objects it was tuned on (angel 0.92\,mm, bell 1.27\,mm, restoration piece 1.09\,mm; memorisation is a possible reason) but fails on unseen objects (\cref{t:base}). \emph{Plain} is DA-V2 alone, \emph{branch} adds the CNN branch; neither has a phase or order head. On all scans the coarse depth of the combined network is worse than the plain baseline (5.73 against 3.96\,mm) and on the unseen objects it is better (9.31 against 12.08\,mm). The phase heads therefore do not improve the coarse depth, and the gain of the combination comes from the matcher. A depth-only network starting from the synthetic model reaches a coarse MAE of 10.49\,mm and 12.8\% within 1\,mm on unseen objects (\cref{t:staged}, P+D separate), the second depth-only baseline in the main paper.

\begin{table}[!t]
\centering\small
\caption{Depth-only baselines: MAE (mm) / share within 1\,mm. Ours: DA-V2, default matcher, 41 scans.}
\label{t:base}
\setlength{\tabcolsep}{2pt}
\begin{tabular}{@{}lrrrr@{}}\toprule
Group & Plain & Branch & Coarse (ours) & Final (ours)\\\midrule
All scans & 3.96/42.1 & 5.04/16.5 & 5.73/13.6 & 5.17/82.7\\
Unseen & 12.08/2.1 & 12.30/1.8 & 9.31/6.8 & 5.07/82.4\\
Figurine held-out & 1.60/45.2 & 1.48/47.7 & 3.74/20.1 & 2.53/90.5\\\bottomrule
\end{tabular}
\end{table}

\subsection{Components and loss terms}
\noindent \Cref{t:comp} changes one component of the real stage (default matcher; all scans / unseen objects). The first row is the final network.
\begin{itemize}
\item \textbf{Phase CNN.} Removing the 132k-parameter dilated phase CNN, so that the phase head reads only the decoder features, costs 0.2\% within 1\,mm (80.8\% against 81.0\%) and leaves the MAE on the unseen objects unchanged (4.93\,mm).
\item \textbf{Fringe branch.} Removing it costs 0.2\% within 1\,mm and 0.7\,mm MAE on all scans (1.0\% and 0.7\,mm on the unseen objects). This is at the limit of what the test set resolves.
\item \textbf{Order loss.} The auxiliary absolute-coordinate loss trades trusted share for trusted precision. Without it 87.2\% of the pixels are trusted (82.7\% with it), but on the unseen objects the trusted pixels have an RMSE90 of 0.83\,mm instead of 0.08\,mm, and the final result there is slightly worse (80.8\% against 82.0\%).
\item \textbf{Phase weight.} 0.5 or 2 gives no clear difference.
\item \textbf{Epochs.} 80 epochs are clearly worse than 200 (75.4\% against 81.0\%).
\end{itemize}

\begin{table}[!t]
\centering\small
\caption{Component ablations of the real stage (default matcher; all 41 scans / unseen objects). Within 1\,mm and MAE (mm). All rows except the first are dilated variants.}
\label{t:comp}
\setlength{\tabcolsep}{2pt}
\begin{tabular}{@{}lrrr@{}}\toprule
Variant & Within & MAE & Trusted\\\midrule
\hl DA-V2, U-Net phase CNN (final) & 82.7/82.4 & 5.17/5.07 & 86.1\%\\
Dilated variant (reference) & 81.0/82.0 & 5.29/4.93 & 82.7\%\\
No phase CNN & 80.8/81.5 & 5.36/4.93 & 82.4\%\\
No fringe branch & 80.8/81.0 & 5.98/5.66 & 83.4\%\\
Order loss off & 83.4/80.8 & 5.14/5.84 & 87.2\%\\
Phase weight 0.5 (default 2) & 79.7/82.2 & 4.77/4.44 & 81.4\%\\
80 real epochs (default 200) & 75.4/80.2 & 6.43/5.43 & 75.6\%\\\bottomrule
\end{tabular}
\end{table}

\subsection{Separate, then joint training}
\noindent A test was done to determine whether the phase and depth parts gain from being trained separately first and then together. All runs use the dilated variant, start from the same synthetic four-carrier model and use the same real data, with 200 epochs in total (tuned matcher, one seed per variant; \cref{t:staged}).
\begin{itemize}
\item \textbf{P} and \textbf{D}: 100 epochs with phase and order losses only, or depth loss only.
\item \textbf{P+D}: phases from P and coarse depth from D, two networks and no joint training.
\item \textbf{J1}, \textbf{J2}: P or D followed by 100 joint epochs at a rate of $10^{-5}$.
\item \textbf{C}: the main model plus 100 joint epochs (300 in total), as a control for training length.
\end{itemize}
No staged variant beats the joint model with the same number of epochs. J1 is equal on the unseen objects (+0.3\%) and 1.3\% worse on all scans, J2 is worse on both. Keeping the networks separate is worse on the unseen objects. The depth-only network has much better coarse depth on the 41 scans, most of which are training objects (MAE 3.94 vs 6.57\,mm), but not on unseen ones (10.49 vs 9.72\,mm). The phase-only network has a slightly larger phase error than the joint model (median 0.203 against 0.187\,rad for the $x_{10.67}$ carrier), but it had half the epochs. The control is the best variant on all scans (+1.2\%) and equal on the unseen objects. Joint training from the start is therefore retained, and a longer schedule may yield a small gain.

\begin{table}[!t]
\centering\small
\caption{Separate and joint training (dilated, tuned matcher). Within; MAE/MAE90 in mm.}
\label{t:staged}
\setlength{\tabcolsep}{2pt}
\begin{tabular}{@{}lrrrr@{}}\toprule
 & \multicolumn{2}{c}{All 41 scans} & \multicolumn{2}{c}{Unseen}\\
Variant & Within & MAE/MAE90 & Within & MAE/MAE90\\\midrule
\hl Main (joint, 200 ep.) & 89.7\% & 1.30/0.07 & 85.9\% & 2.53/0.33\\
P+D separate & 88.3\% & 1.61/0.11 & 82.1\% & 4.38/1.13\\
J1 (P then joint) & 88.4\% & 1.61/0.11 & 86.2\% & 2.34/0.28\\
J2 (D then joint) & 83.5\% & 2.00/0.35 & 81.6\% & 3.81/0.94\\
C (control, 300 ep.) & 90.9\% & 1.21/0.06 & 85.8\% & 2.60/0.35\\\bottomrule
\end{tabular}
\end{table}

\section{Ray-wise phase matching}\label{s:match}
\gist{The four carriers together fix the fringe order only if the coarse depth is within about $\pm33$\,mm. The tuned settings gain about 4\% over the default on unseen objects, and no single carrier or carrier subset reaches the joint result.}

\subsection{Cost along a ray}
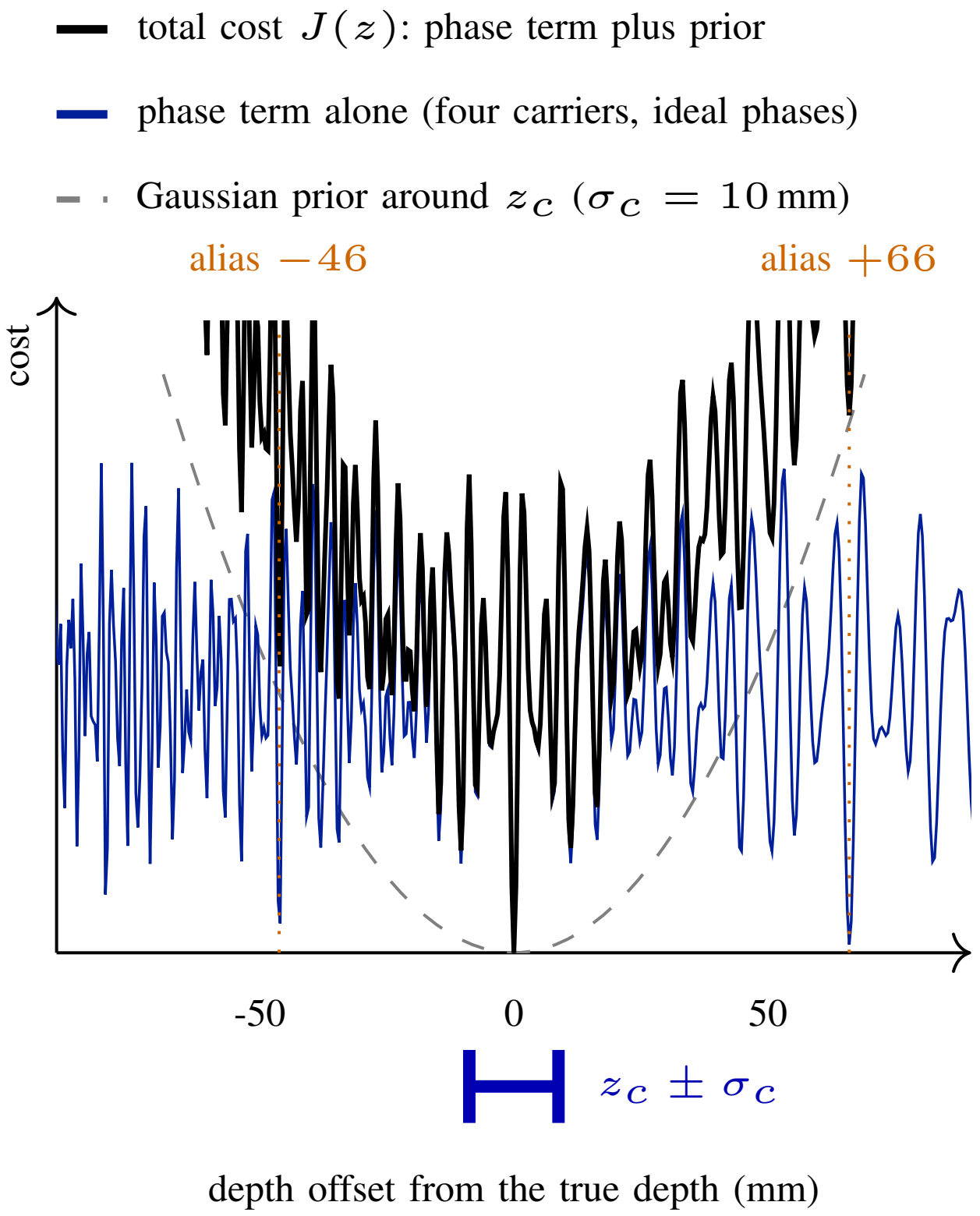
\begin{figure}[!t]
\centering
\resizebox{0.95\columnwidth}{!}{\begin{tikzpicture}[font=\scriptsize]
% J(z) of Eq. (3) along one real camera ray of the calibrated rig (pixel 417.6, 365.8 of the 1024x832 view, true depth 0.3494 m), ideal phases, phase noise 0.6 rad, prior 10 mm centred on the true depth
\begin{scope}[xscale=0.0230,yscale=0.11]
\draw[->] (-90,0)--(90,0);
\draw[->] (-90,0)--(-90,27) node[left,font=\tiny,rotate=90,anchor=south east] {cost};
\begin{scope}\clip (-90,0) rectangle (90,26);
\draw[gray,dashed,domain=-69:69,samples=60] plot (\x,{0.5*(\x/10)^2});
\draw[blue!80!cyan!60!black,line width=0.35pt] plot coordinates {(-100.0,9.13) (-99.6,4.52) (-99.2,14.19) (-98.8,12.82) (-98.4,7.18) (-98.0,11.25) (-97.6,13.67) (-97.2,14.14) (-96.8,12.28) (-96.4,3.83) (-96.0,6.27) (-95.6,18.62) (-95.2,16.36) (-94.8,7.85) (-94.4,9.28) (-94.0,9.90) (-93.6,9.28) (-93.2,13.45) (-92.8,12.59) (-92.4,9.92) (-92.0,13.88) (-91.6,12.13) (-91.2,5.18) (-90.8,9.02) (-90.4,15.16) (-90.0,12.42) (-89.6,11.86) (-89.2,13.53) (-88.8,8.24) (-88.4,5.96) (-88.0,11.82) (-87.6,13.70) (-87.2,12.53) (-86.8,14.56) (-86.4,11.20) (-86.0,4.38) (-85.6,8.40) (-85.2,16.02) (-84.8,13.05) (-84.4,8.92) (-84.0,12.26) (-83.6,13.20) (-83.2,9.74) (-82.8,9.52) (-82.4,9.38) (-82.0,7.87) (-81.6,13.38) (-81.2,20.14) (-80.8,13.63) (-80.4,2.40) (-80.0,4.31) (-79.6,13.51) (-79.2,15.76) (-78.8,13.50) (-78.4,11.98) (-78.0,8.83) (-77.6,7.68) (-77.2,12.92) (-76.8,15.01) (-76.4,7.76) (-76.0,4.40) (-75.6,13.21) (-75.2,20.16) (-74.8,14.69) (-74.4,6.58) (-74.0,5.16) (-73.6,7.28) (-73.2,11.03) (-72.8,17.12) (-72.4,18.38) (-72.0,10.22) (-71.6,3.66) (-71.2,8.42) (-70.8,15.23) (-70.4,12.85) (-70.0,7.62) (-69.6,9.33) (-69.2,13.85) (-68.8,14.00) (-68.4,12.39) (-68.0,11.94) (-67.6,9.02) (-67.2,4.67) (-66.8,7.09) (-66.4,15.81) (-66.0,19.11) (-65.6,12.96) (-65.2,7.40) (-64.8,8.76) (-64.4,10.83) (-64.0,9.08) (-63.6,8.76) (-63.2,12.85) (-62.8,15.26) (-62.4,12.44) (-62.0,9.97) (-61.6,11.55) (-61.2,12.07) (-60.8,8.34) (-60.4,6.36) (-60.0,10.65) (-59.6,15.35) (-59.2,13.91) (-58.8,10.16) (-58.4,10.93) (-58.0,13.86) (-57.6,12.02) (-57.2,6.66) (-56.8,5.57) (-56.4,10.34) (-56.0,14.54) (-55.6,14.55) (-55.2,13.64) (-54.8,13.77) (-54.4,11.55) (-54.0,6.13) (-53.6,3.76) (-53.2,9.08) (-52.8,16.54) (-52.4,17.24) (-52.0,11.39) (-51.6,7.47) (-51.2,9.79) (-50.8,13.47) (-50.4,13.08) (-50.0,10.16) (-49.6,8.92) (-49.2,9.27) (-48.8,8.84) (-48.4,8.95) (-48.0,12.90) (-47.6,18.64) (-47.2,18.95) (-46.8,11.08) (-46.4,2.12) (-46.0,1.21) (-45.6,8.60) (-45.2,16.08) (-44.8,17.44) (-44.4,14.40) (-44.0,11.70) (-43.6,10.42) (-43.2,8.90) (-42.8,7.75) (-42.4,9.50) (-42.0,13.50) (-41.6,14.88) (-41.2,10.85) (-40.8,5.60) (-40.4,5.99) (-40.0,12.85) (-39.6,19.29) (-39.2,18.62) (-38.8,11.80) (-38.4,5.53) (-38.0,4.33) (-37.6,6.82) (-37.2,9.63) (-36.8,12.06) (-36.4,15.13) (-36.0,17.72) (-35.6,16.55) (-35.2,10.89) (-34.8,5.09) (-34.4,4.54) (-34.0,9.48) (-33.6,14.46) (-33.2,14.47) (-32.8,10.49) (-32.4,7.81) (-32.0,9.58) (-31.6,13.52) (-31.2,15.21) (-30.8,13.41) (-30.4,10.82) (-30.0,10.01) (-29.6,10.27) (-29.2,9.32) (-28.8,7.20) (-28.4,6.82) (-28.0,10.30) (-27.6,15.64) (-27.2,18.20) (-26.8,15.64) (-26.4,10.49) (-26.0,7.42) (-25.6,8.27) (-25.2,10.55) (-24.8,10.84) (-24.4,8.88) (-24.0,7.72) (-23.6,9.87) (-23.2,14.13) (-22.8,16.72) (-22.4,15.36) (-22.0,11.60) (-21.6,8.92) (-21.2,9.01) (-20.8,10.30) (-20.4,10.31) (-20.0,8.81) (-19.6,8.02) (-19.2,9.89) (-18.8,13.46) (-18.4,15.57) (-18.0,14.20) (-17.6,10.72) (-17.2,8.64) (-16.8,10.00) (-16.4,13.20) (-16.0,14.58) (-15.6,12.11) (-15.2,7.50) (-14.8,4.61) (-14.4,5.92) (-14.0,10.38) (-13.6,14.71) (-13.2,16.43) (-12.8,15.64) (-12.4,14.11) (-12.0,12.95) (-11.6,11.56) (-11.2,8.90) (-10.8,5.48) (-10.4,3.67) (-10.0,5.71) (-9.6,11.29) (-9.2,17.11) (-8.8,19.29) (-8.4,16.43) (-8.0,10.84) (-7.6,6.62) (-7.2,6.47) (-6.8,9.70) (-6.4,13.26) (-6.0,14.41) (-5.6,12.79) (-5.2,10.16) (-4.8,8.53) (-4.4,8.45) (-4.0,9.03) (-3.6,9.37) (-3.2,9.78) (-2.8,11.36) (-2.4,14.54) (-2.0,17.93) (-1.6,18.93) (-1.2,15.76) (-0.8,9.23) (-0.4,2.71) (-0.0,0.00) (0.4,2.70) (0.8,9.15) (1.2,15.63) (1.6,18.89) (2.0,18.10) (2.4,14.88) (2.8,11.68) (3.2,9.93) (3.6,9.42) (4.0,9.15) (4.4,8.61) (4.8,8.35) (5.2,9.27) (5.6,11.47) (6.0,13.75) (6.4,14.36) (6.8,12.43) (7.2,8.94) (7.6,6.29) (8.0,6.65) (8.4,10.34) (8.8,15.44) (9.2,18.92) (9.6,18.62) (10.0,14.62) (10.4,9.10) (10.8,4.88) (11.2,3.69) (11.6,5.36) (12.0,8.30) (12.4,10.87) (12.8,12.41) (13.2,13.36) (13.6,14.45) (14.0,15.74) (14.4,16.44) (14.8,15.44) (15.2,12.44) (15.6,8.42) (16.0,5.27) (16.4,4.65) (16.8,6.87) (17.2,10.62) (17.6,13.73) (18.0,14.56) (18.4,12.99) (18.8,10.39) (19.2,8.72) (19.6,9.19) (20.0,11.54) (20.4,14.24) (20.8,15.58) (21.2,14.74) (21.6,12.26) (22.0,9.58) (22.4,8.11) (22.8,8.28) (23.2,9.45) (23.6,10.43) (24.0,10.40) (24.4,9.50) (24.8,8.70) (25.2,9.08) (25.6,11.03) (26.0,13.86) (26.4,16.18) (26.8,16.69) (27.2,15.06) (27.6,12.06) (28.0,9.20) (28.4,7.77) (28.8,8.14) (29.2,9.61) (29.6,10.87) (30.0,10.92) (30.4,9.66) (30.8,8.01) (31.2,7.30) (31.6,8.48) (32.0,11.44) (32.4,15.03) (32.8,17.65) (33.2,18.06) (33.6,16.07) (34.0,12.59) (34.4,9.12) (34.8,6.97) (35.2,6.63) (35.6,7.67) (36.0,9.11) (36.4,10.07) (36.8,10.25) (37.2,10.04) (37.6,10.15) (38.0,11.10) (38.4,12.77) (38.8,14.46) (39.2,15.21) (39.6,14.45) (40.0,12.34) (40.4,9.79) (40.8,8.05) (41.2,7.99) (41.6,9.71) (42.0,12.37) (42.4,14.58) (42.8,15.12) (43.2,13.50) (43.6,10.26) (44.0,6.70) (44.4,4.34) (44.8,4.19) (45.2,6.36) (45.6,10.04) (46.0,13.92) (46.4,16.75) (46.8,17.87) (47.2,17.33) (47.6,15.74) (48.0,13.83) (48.4,12.12) (48.8,10.70) (49.2,9.31) (49.6,7.74) (50.0,6.01) (50.4,4.58) (50.8,4.18) (51.2,5.39) (51.6,8.34) (52.0,12.45) (52.4,16.57) (52.8,19.38) (53.2,19.91) (53.6,17.93) (54.0,14.10) (54.4,9.73) (54.8,6.28) (55.2,4.83) (55.6,5.65) (56.0,8.21) (56.4,11.37) (56.8,13.93) (57.2,15.03) (57.6,14.47) (58.0,12.68) (58.4,10.49) (58.8,8.70) (59.2,7.82) (59.6,7.89) (60.0,8.59) (60.4,9.48) (60.8,10.23) (61.2,10.83) (61.6,11.49) (62.0,12.51) (62.4,14.01) (62.8,15.77) (63.2,17.23) (63.6,17.68) (64.0,16.56) (64.4,13.73) (64.8,9.65) (65.2,5.28) (65.6,1.84) (66.0,0.34) (66.4,1.31) (66.8,4.55) (67.2,9.20) (67.6,14.03) (68.0,17.81) (68.4,19.72) (68.8,19.53) (69.2,17.63) (69.6,14.82) (70.0,12.01) (70.4,9.91) (70.8,8.80) (71.2,8.59) (71.6,8.86) (72.0,9.19) (72.4,9.28) (72.8,9.12) (73.2,8.94) (73.6,9.05) (74.0,9.68) (74.4,10.84) (74.8,12.24) (75.2,13.41) (75.6,13.88) (76.0,13.37) (76.4,11.96) (76.8,10.07) (77.2,8.35) (77.6,7.47) (78.0,7.89) (78.4,9.62) (78.8,12.28) (79.2,15.11) (79.6,17.27) (80.0,18.05) (80.4,17.14) (80.8,14.70) (81.2,11.33) (81.6,7.86) (82.0,5.14) (82.4,3.76) (82.8,3.91) (83.2,5.37) (83.6,7.62) (84.0,10.00) (84.4,11.97) (84.8,13.22) (85.2,13.74) (85.6,13.76) (86.0,13.65) (86.4,13.69) (86.8,14.02) (87.2,14.53) (87.6,14.92) (88.0,14.85) (88.4,14.06) (88.8,12.49) (89.2,10.36) (89.6,8.09) (90.0,6.22) (90.4,5.22) (90.8,5.34) (91.2,6.57) (91.6,8.56) (92.0,10.79) (92.4,12.69) (92.8,13.81) (93.2,13.94) (93.6,13.20) (94.0,11.94) (94.4,10.68) (94.8,9.90) (95.2,9.91) (95.6,10.77) (96.0,12.23) (96.4,13.85) (96.8,15.12) (97.2,15.60) (97.6,15.09) (98.0,13.62) (98.4,11.53) (98.8,9.31) (99.2,7.46) (99.6,6.41) (100.0,6.34)};
\draw[line width=0.6pt] plot coordinates {(-100.0,59.13) (-99.6,54.12) (-99.2,63.40) (-98.8,61.63) (-98.4,55.59) (-98.0,59.27) (-97.6,61.30) (-97.2,61.38) (-96.8,59.13) (-96.4,50.29) (-96.0,52.35) (-95.6,64.32) (-95.2,61.68) (-94.8,52.78) (-94.4,53.83) (-94.0,54.08) (-93.6,53.08) (-93.2,56.88) (-92.8,55.65) (-92.4,52.61) (-92.0,56.20) (-91.6,54.09) (-91.2,46.77) (-90.8,50.24) (-90.4,56.02) (-90.0,52.92) (-89.6,52.00) (-89.2,53.32) (-88.8,47.67) (-88.4,45.03) (-88.0,50.54) (-87.6,52.07) (-87.2,50.55) (-86.8,52.23) (-86.4,48.53) (-86.0,41.36) (-85.6,45.03) (-85.2,52.31) (-84.8,49.01) (-84.4,44.54) (-84.0,47.54) (-83.6,48.14) (-83.2,44.35) (-82.8,43.80) (-82.4,43.33) (-82.0,41.49) (-81.6,46.67) (-81.2,53.11) (-80.8,46.28) (-80.4,34.73) (-80.0,36.31) (-79.6,45.19) (-79.2,47.12) (-78.8,44.55) (-78.4,42.71) (-78.0,39.25) (-77.6,37.79) (-77.2,42.72) (-76.8,44.50) (-76.4,36.94) (-76.0,33.28) (-75.6,41.78) (-75.2,48.44) (-74.8,42.67) (-74.4,34.26) (-74.0,32.54) (-73.6,34.36) (-73.2,37.82) (-72.8,43.62) (-72.4,44.59) (-72.0,36.14) (-71.6,29.29) (-71.2,33.77) (-70.8,40.29) (-70.4,37.64) (-70.0,32.12) (-69.6,33.55) (-69.2,37.79) (-68.8,37.67) (-68.4,35.78) (-68.0,35.06) (-67.6,31.87) (-67.2,27.25) (-66.8,29.40) (-66.4,37.86) (-66.0,40.89) (-65.6,34.48) (-65.2,28.66) (-64.8,29.76) (-64.4,31.57) (-64.0,29.56) (-63.6,28.98) (-63.2,32.82) (-62.8,34.98) (-62.4,31.90) (-62.0,29.19) (-61.6,30.52) (-61.2,30.79) (-60.8,26.82) (-60.4,24.60) (-60.0,28.65) (-59.6,33.11) (-59.2,31.43) (-58.8,27.44) (-58.4,27.98) (-58.0,30.68) (-57.6,28.61) (-57.2,23.02) (-56.8,21.70) (-56.4,26.24) (-56.0,30.22) (-55.6,30.01) (-55.2,28.88) (-54.8,28.78) (-54.4,26.35) (-54.0,20.71) (-53.6,18.13) (-53.2,23.23) (-52.8,30.48) (-52.4,30.97) (-52.0,24.91) (-51.6,20.79) (-51.2,22.90) (-50.8,26.37) (-50.4,25.78) (-50.0,22.66) (-49.6,21.22) (-49.2,21.38) (-48.8,20.75) (-48.4,20.67) (-48.0,24.42) (-47.6,29.97) (-47.2,30.09) (-46.8,22.03) (-46.4,12.88) (-46.0,11.79) (-45.6,19.00) (-45.2,26.30) (-44.8,27.47) (-44.4,24.26) (-44.0,21.38) (-43.6,19.92) (-43.2,18.23) (-42.8,16.91) (-42.4,18.49) (-42.0,22.32) (-41.6,23.53) (-41.2,19.34) (-40.8,13.92) (-40.4,14.15) (-40.0,20.85) (-39.6,27.13) (-39.2,26.30) (-38.8,19.32) (-38.4,12.90) (-38.0,11.55) (-37.6,13.89) (-37.2,16.55) (-36.8,18.83) (-36.4,21.76) (-36.0,24.20) (-35.6,22.89) (-35.2,17.09) (-34.8,11.15) (-34.4,10.46) (-34.0,15.26) (-33.6,20.10) (-33.2,19.98) (-32.8,15.87) (-32.4,13.06) (-32.0,14.70) (-31.6,18.51) (-31.2,20.07) (-30.8,18.15) (-30.4,15.44) (-30.0,14.51) (-29.6,14.65) (-29.2,13.59) (-28.8,11.34) (-28.4,10.85) (-28.0,14.22) (-27.6,19.45) (-27.2,21.90) (-26.8,19.23) (-26.4,13.98) (-26.0,10.80) (-25.6,11.54) (-25.2,13.73) (-24.8,13.92) (-24.4,11.85) (-24.0,10.60) (-23.6,12.66) (-23.2,16.82) (-22.8,19.32) (-22.4,17.87) (-22.0,14.02) (-21.6,11.25) (-21.2,11.25) (-20.8,12.46) (-20.4,12.40) (-20.0,10.81) (-19.6,9.94) (-19.2,11.73) (-18.8,15.22) (-18.4,17.27) (-18.0,15.82) (-17.6,12.27) (-17.2,10.12) (-16.8,11.41) (-16.4,14.54) (-16.0,15.86) (-15.6,13.32) (-15.2,8.65) (-14.8,5.71) (-14.4,6.96) (-14.0,11.36) (-13.6,15.64) (-13.2,17.30) (-12.8,16.46) (-12.4,14.88) (-12.0,13.67) (-11.6,12.23) (-11.2,9.52) (-10.8,6.06) (-10.4,4.21) (-10.0,6.21) (-9.6,11.75) (-9.2,17.54) (-8.8,19.68) (-8.4,16.78) (-8.0,11.16) (-7.6,6.91) (-7.2,6.73) (-6.8,9.93) (-6.4,13.47) (-6.0,14.59) (-5.6,12.94) (-5.2,10.29) (-4.8,8.65) (-4.4,8.55) (-4.0,9.11) (-3.6,9.44) (-3.2,9.83) (-2.8,11.40) (-2.4,14.57) (-2.0,17.95) (-1.6,18.95) (-1.2,15.76) (-0.8,9.23) (-0.4,2.71) (-0.0,0.00) (0.4,2.70) (0.8,9.16) (1.2,15.64) (1.6,18.90) (2.0,18.12) (2.4,14.91) (2.8,11.72) (3.2,9.98) (3.6,9.48) (4.0,9.23) (4.4,8.71) (4.8,8.47) (5.2,9.41) (5.6,11.63) (6.0,13.93) (6.4,14.56) (6.8,12.66) (7.2,9.20) (7.6,6.58) (8.0,6.97) (8.4,10.69) (8.8,15.82) (9.2,19.34) (9.6,19.08) (10.0,15.12) (10.4,9.64) (10.8,5.46) (11.2,4.32) (11.6,6.03) (12.0,9.02) (12.4,11.64) (12.8,13.23) (13.2,14.23) (13.6,15.37) (14.0,16.72) (14.4,17.47) (14.8,16.54) (15.2,13.60) (15.6,9.63) (16.0,6.55) (16.4,6.00) (16.8,8.29) (17.2,12.10) (17.6,15.28) (18.0,16.18) (18.4,14.68) (18.8,12.16) (19.2,10.57) (19.6,11.12) (20.0,13.54) (20.4,16.32) (20.8,17.75) (21.2,16.99) (21.6,14.59) (22.0,12.00) (22.4,10.62) (22.8,10.88) (23.2,12.14) (23.6,13.21) (24.0,13.28) (24.4,12.47) (24.8,11.77) (25.2,12.26) (25.6,14.31) (26.0,17.24) (26.4,19.66) (26.8,20.29) (27.2,18.76) (27.6,15.87) (28.0,13.12) (28.4,11.80) (28.8,12.29) (29.2,13.87) (29.6,15.25) (30.0,15.42) (30.4,14.28) (30.8,12.75) (31.2,12.17) (31.6,13.47) (32.0,16.56) (32.4,20.28) (32.8,23.03) (33.2,23.57) (33.6,21.72) (34.0,18.37) (34.4,15.04) (34.8,13.02) (35.2,12.82) (35.6,14.00) (36.0,15.59) (36.4,16.70) (36.8,17.02) (37.2,16.96) (37.6,17.22) (38.0,18.32) (38.4,20.15) (38.8,21.99) (39.2,22.90) (39.6,22.29) (40.0,20.34) (40.4,17.95) (40.8,16.37) (41.2,16.48) (41.6,18.37) (42.0,21.19) (42.4,23.57) (42.8,24.28) (43.2,22.83) (43.6,19.76) (44.0,16.38) (44.4,14.20) (44.8,14.22) (45.2,16.57) (45.6,20.43) (46.0,24.50) (46.4,27.52) (46.8,28.83) (47.2,28.47) (47.6,27.06) (48.0,25.35) (48.4,23.84) (48.8,22.60) (49.2,21.42) (49.6,20.04) (50.0,18.51) (50.4,17.28) (50.8,17.08) (51.2,18.50) (51.6,21.65) (52.0,25.97) (52.4,30.30) (52.8,33.32) (53.2,34.06) (53.6,32.29) (54.0,28.68) (54.4,24.53) (54.8,21.30) (55.2,20.06) (55.6,21.11) (56.0,23.89) (56.4,27.27) (56.8,30.06) (57.2,31.39) (57.6,31.06) (58.0,29.50) (58.4,27.54) (58.8,25.99) (59.2,25.34) (59.6,25.65) (60.0,26.59) (60.4,27.72) (60.8,28.72) (61.2,29.56) (61.6,30.46) (62.0,31.73) (62.4,33.48) (62.8,35.49) (63.2,37.20) (63.6,37.91) (64.0,37.04) (64.4,34.46) (64.8,30.64) (65.2,26.54) (65.6,23.36) (66.0,22.12) (66.4,23.36) (66.8,26.86) (67.2,31.78) (67.6,36.87) (68.0,40.93) (68.4,43.11) (68.8,43.20) (69.2,41.57) (69.6,39.04) (70.0,36.51) (70.4,34.69) (70.8,33.87) (71.2,33.93) (71.6,34.50) (72.0,35.11) (72.4,35.49) (72.8,35.62) (73.2,35.73) (73.6,36.13) (74.0,37.06) (74.4,38.52) (74.8,40.21) (75.2,41.68) (75.6,42.45) (76.0,42.25) (76.4,41.14) (76.8,39.56) (77.2,38.15) (77.6,37.58) (78.0,38.31) (78.4,40.36) (78.8,43.33) (79.2,46.47) (79.6,48.95) (80.0,50.05) (80.4,49.46) (80.8,47.34) (81.2,44.29) (81.6,41.15) (82.0,38.76) (82.4,37.71) (82.8,38.19) (83.2,39.98) (83.6,42.56) (84.0,45.28) (84.4,47.59) (84.8,49.18) (85.2,50.03) (85.6,50.40) (86.0,50.63) (86.4,51.02) (86.8,51.69) (87.2,52.55) (87.6,53.29) (88.0,53.57) (88.4,53.13) (88.8,51.92) (89.2,50.14) (89.6,48.23) (90.0,46.72) (90.4,46.08) (90.8,46.57) (91.2,48.16) (91.6,50.51) (92.0,53.11) (92.4,55.38) (92.8,56.87) (93.2,57.37) (93.6,57.00) (94.0,56.12) (94.4,55.23) (94.8,54.83) (95.2,55.23) (95.6,56.46) (96.0,58.31) (96.4,60.31) (96.8,61.97) (97.2,62.84) (97.6,62.71) (98.0,61.64) (98.4,59.94) (98.8,58.11) (99.2,56.67) (99.6,56.01) (100.0,56.34)};
\end{scope}
\foreach \a in {-46.2,66.0}{\draw[orange!80!black,dotted] (\a,0)--(\a,26);}
\draw[black,line width=1.1pt] (-90,38)--(-80,38) node[right,font=\tiny,text=black] {total cost $J(z)$: phase term plus prior};
\draw[blue!80!cyan!60!black,line width=1.1pt] (-90,34.5)--(-80,34.5) node[right,font=\tiny,text=black] {phase term alone (four carriers, ideal phases)};
\draw[gray,dashed,line width=0.8pt] (-90,31)--(-80,31) node[right,font=\tiny,text=black] {Gaussian prior around $z_c$ ($\sigma_c=10$\,mm)};
\node[font=\tiny,orange!80!black] at (-46.2,28.5) {alias $-46$};
\node[font=\tiny,orange!80!black] at (66,28.5) {alias $+66$};
\foreach \a in {-50,0,50}{\node[below,font=\tiny] at (\a,-0.8) {\a};}
\draw[blue!70!black,line width=1.6pt,{Bar}-{Bar}] (-10,-5.5)--(10,-5.5) node[right,font=\tiny] {$z_c\pm\sigma_c$};
\node[font=\tiny,anchor=north] at (0,-8) {depth offset from the true depth (mm)};
\end{scope}
\end{tikzpicture}}
\caption{Cost $J(z)$ along one real camera ray of the calibrated rig, computed with ideal phases (pixel 418, 366 of the $1024{\times}832$ view, true depth 0.349\,m; pinhole projector model, phase noise 0.6\,rad, prior 10\,mm centred on the true depth $z_c$). The horizontal axis is the depth offset from the true depth; the vertical axis is cost in the units of Eq.~(3), clipped at 26. \emph{Blue:} phase term of the four carriers alone. It is a sum of four fast oscillations (depth periods of roughly 4--7\,mm per carrier), so it has many shallow minima; it reaches 0 at the true depth and almost 0 at the two aliases ($-46$ and $+66$\,mm, cost 0.37 and 0.34), where all four carriers nearly agree again. \emph{Dashed grey:} Gaussian prior around the coarse depth. \emph{Black:} their sum. The prior adds about 22 at the $+66$\,mm alias, so the true minimum is the deepest by a wide margin as long as the coarse depth is within roughly half the alias spacing; a coarse error of $\pm\sigma_c$ (blue bar) does not change that. The aliases are not exactly periodic and differ between rays.}
\label{f:cost}
\end{figure}

\noindent \Cref{f:cost} shows the cost $J(z)$ of Eq.~(3) of the main paper along one real camera ray of the calibrated rig, computed with ideal phases, equal carrier weights, phase noise 0.6\,rad and a 10\,mm prior centred on the true depth. The phase term alone has near-minima wherever all four carriers nearly agree again; they are not exactly periodic. On this ray the nearest are at $-46$ and $+66$\,mm and cost 0.37 and 0.34, below the cost of a single carrier being one phase-noise unit off (0.49). The Gaussian prior around the coarse depth $z_c$ makes the nearest minimum the deepest. The margin to the second-best minimum is the alternative trust measure of \cref{t:trust}.

\noindent\textbf{Spacing of the minima.} The 66 and 15\,mm of the main paper are medians of the distance from the true depth to the nearest depth at which all carriers agree within 0.1 period, measured on the real scanner (angel scans, median depth 0.35\,m). They are 66.0\,mm for the four carriers (10th percentile 29.5\,mm) and 15.0\,mm for the two 16-pixel carriers alone (10th percentile 10.0\,mm). The ray of \cref{f:cost} was chosen to have exactly these two values; over 300 random pixels at 0.30--0.40\,m the medians are 74.5 and 13.8\,mm. Both are therefore typical, not guaranteed, spacings: a coarse error of about half the spacing is tolerated at a typical pixel, and less at pixels with a nearer alias.

\subsection{Matcher settings}\label{s:settings}
\noindent\textbf{Grid.} The tuned settings were selected on the four tuning objects only, from the $3{\times}3{\times}2$ grid of \cref{t:grid} (dilated variant). The prior width gives most of the gain: with a prior of 10\,mm and the other two settings at default, the unseen objects already reach 86.6--87.0\%. A narrow prior needs a coarse depth that is correct within a few centimetres and does not help where it is far off (vase, \cref{s:vase}). On the tuning objects the rows differ by up to 10\% and the selected row is the best of them. On the unseen objects, which were not used for the selection, nine rows lie between 86.4\% and 87.0\%, and the selected row (85.9\%) is 1.1\% below the best of them, well inside the interval over scans (\cref{t:sets}). With the final network the selected setting gives 86.7\% and the default 82.4\% within 1\,mm on the unseen objects.

\begin{table}[!t]
\centering\small
\caption{Matcher grid (dilated variant). Within = share within 1\,mm; MAE in mm; Trusted = trusted share of the evaluated pixels; DZ = edge dead zone. Tuning = the four tuning objects; Unseen = fossil and benchmark. Shaded: the default (first version) and the selected setting.}
\label{t:grid}
\setlength{\tabcolsep}{2pt}\footnotesize
\begin{tabular}{@{}rrlrrrrr@{}}\toprule
 & & & \multicolumn{2}{c}{Tuning} & \multicolumn{3}{c}{Unseen}\\
Prior (mm) & Phase (rad) & DZ & Within & MAE & Within & MAE & Trusted\\\midrule
10 & 0.3 & off & 89.5\% & 1.32 & 87.0\% & 2.37 & 89.0\%\\
10 & 0.3 & on & 88.6\% & 1.64 & 86.6\% & 2.59 & 88.2\%\\
10 & 0.45 & off & 90.7\% & 0.96 & 86.8\% & 2.33 & 89.3\%\\
10 & 0.45 & on & 89.7\% & 1.24 & 86.4\% & 2.54 & 88.4\%\\
\hl 10 & 0.6 & off & 91.0\% & 0.89 & 85.9\% & 2.53 & 89.3\%\\
10 & 0.6 & on & 89.9\% & 1.17 & 85.5\% & 2.72 & 88.3\%\\
\midrule
15 & 0.3 & off & 87.9\% & 2.00 & 85.9\% & 2.81 & 88.4\%\\
15 & 0.3 & on & 86.5\% & 2.51 & 85.4\% & 3.13 & 87.4\%\\
15 & 0.45 & off & 89.8\% & 1.30 & 86.9\% & 2.33 & 89.4\%\\
15 & 0.45 & on & 88.7\% & 1.65 & 86.5\% & 2.57 & 88.4\%\\
15 & 0.6 & off & 90.7\% & 1.00 & 86.8\% & 2.32 & 89.5\%\\
15 & 0.6 & on & 89.7\% & 1.29 & 86.4\% & 2.54 & 88.5\%\\
\midrule
25 & 0.3 & off & 82.6\% & 4.47 & 82.9\% & 4.28 & 87.3\%\\
\hl 25 & 0.3 & on & 80.6\% & 5.41 & 82.0\% & 4.93 & 85.8\%\\
25 & 0.45 & off & 87.4\% & 2.29 & 85.3\% & 3.08 & 88.5\%\\
25 & 0.45 & on & 85.8\% & 2.92 & 84.7\% & 3.48 & 87.3\%\\
25 & 0.6 & off & 89.1\% & 1.57 & 86.5\% & 2.45 & 89.3\%\\
25 & 0.6 & on & 87.8\% & 2.01 & 86.1\% & 2.75 & 88.2\%\\
\bottomrule
\end{tabular}
\end{table}

\noindent\textbf{Wider grid.} \Cref{t:grid2} separates the prior width from the search window and adds phase noise 0.9 and 1.2\,rad (dead zone off, window 40\,mm). The window is $\min(4\sigma_c,\text{window})$, so it matters only for $\sigma_c>5$\,mm. Selected on the tuning objects only: prior 15\,mm, window 40\,mm, phase noise 1.2\,rad. The tuning objects stay within 0.6\% of each other (90.7--91.2\%) for every prior of 8\,mm or more, so they do not rank these settings. On the unseen objects the selected values reach 84.1\% and the original selection 85.9\%, a difference inside the interval over scans (\cref{t:sets}). Priors of 3 and 5\,mm fail on the unseen objects (63--78\%) because the coarse depth error (standard deviation 7.4\,mm, robust 6.2\,mm) is larger than the prior. A window of 60 or 100\,mm gives the same result as 40\,mm.

\begin{table}[!t]
\centering\small
\caption{Wider matcher grid (dilated variant, window 40\,mm, dead zone off). Within 1\,mm / MAE (mm) on the tuning and the unseen objects.}
\label{t:grid2}
\setlength{\tabcolsep}{2pt}
\begin{tabular}{@{}rrrr@{}}\toprule
Prior (mm) & Phase (rad) & Tuning & Unseen\\\midrule
3 & 0.6 & 89.9\% / 1.05 & 62.9\% / 6.04\\
5 & 0.6 & 90.7\% / 0.93 & 77.5\% / 4.03\\
8 & 0.6 & 91.0\% / 0.88 & 84.7\% / 2.78\\
\hl 10 & 0.6 & 91.0\% / 0.89 & 85.9\% / 2.53\\
10 & 1.2 & 90.8\% / 0.91 & 75.3\% / 4.29\\
15 & 0.6 & 90.7\% / 1.00 & 86.8\% / 2.32\\
15 & 0.9 & 91.1\% / 0.87 & 85.7\% / 2.55\\
15 & 1.2 (selected) & 91.2\% / 0.86 & 84.1\% / 2.91\\\bottomrule
\end{tabular}
\end{table}

\noindent\textbf{Closed-form estimates.} The two noise parameters can be read from the labelled tuning scans without a search (\cref{t:sigma}). The scatter of the coarse depth error is 7.4\,mm (standard deviation) or 6.2\,mm (robust, 1.4826 times the median absolute deviation). That of the network phase against the ideal phase is 0.47\,rad, or 0.26\,rad robust, averaged over the carriers (per carrier, robust: $x_{10}$ 0.29, $x_{16}$ 0.20, $y_{10}$ 0.34, $y_{16}$ 0.22\,rad; the 16-pixel carriers are cleaner). The robust estimates, used as prior width and phase noise with a window of 25\,mm, are as good as the selected settings on unseen objects. Halving them loses 5\% and doubling them changes nothing. The estimates are therefore a usable starting point on a new scanner, with a tendency to the wide side.

\begin{table}[!t]
\centering\small
\caption{Matcher settings from the closed-form estimates against the selected ones (dilated variant, unseen objects).}
\label{t:sigma}
\setlength{\tabcolsep}{2pt}
\begin{tabular}{@{}lrrrr@{}}\toprule
Setting & Prior (mm) & Phase (rad) & Within & MAE\\\midrule
robust estimate & 6.2 & 0.26 & 86.7\% & 2.38\\
estimate $\times0.5$ & 3.1 & 0.13 & 81.3\% & 3.53\\
estimate $\times2$ & 12.5 & 0.53 & 86.8\% & 2.32\\
\hl selected & 10.0 & 0.60 & 85.9\% & 2.53\\\bottomrule
\end{tabular}
\end{table}

\subsection{Joint matching against single carriers}\label{s:single}
\noindent \Cref{t:carr} gives the comparison of the main text in full. A single carrier with the coarse depth does not recover the fringe order: the share within 1\,mm rises only from 6.8\% to 10--20\% on the unseen objects, and the share above 5\,mm stays at 64--76\% (67\% for the coarse depth). The joint matching without clean-up already gives 85.2\%.

\begin{table}[!t]
\centering\small
\caption{Single-carrier unwrapping against the joint matching (DA-V2, tuned settings). Within 1\,mm / MAE (mm) / error above 5\,mm.}
\label{t:carr}
\setlength{\tabcolsep}{2pt}
\begin{tabular}{@{}lrr@{}}\toprule
Method & Unseen & All 41 scans\\\midrule
network coarse depth & 6.8\% / 9.31 / 67\% & 13.6\% / 5.73 / 43\%\\
carrier $x_{10}$ alone & 9.9\% / 9.45 / 69\% & 18.0\% / 6.02 / 46\%\\
carrier $x_{16}$ alone & 15.0\% / 9.35 / 64\% & 28.1\% / 5.79 / 44\%\\
carrier $y_{10}$ alone & 13.1\% / 9.38 / 64\% & 24.5\% / 5.88 / 44\%\\
carrier $y_{16}$ alone & 19.6\% / 9.28 / 76\% & 37.0\% / 5.63 / 54\%\\
median of the four & 12.0\% / 9.28 / 68\% & 23.1\% / 5.65 / 46\%\\
four carriers, no clean-up & 85.2\% / 3.18 / 14\% & 90.0\% / 1.63 / 8.5\%\\
\hl four carriers, full & 86.7\% / 2.47 / 11.9\% & 91.7\% / 1.15 / 6.4\%\\\bottomrule
\end{tabular}
\end{table}

\noindent\textbf{Carrier subsets.} \Cref{t:subset} removes carriers from the joint matching (unseen objects). Every subset is worse than the four carriers: the two 16-pixel carriers alone lose 28\%, three carriers lose 10 to 19\%, and the two short-period carriers alone lose 44\%.

\begin{table}[!t]
\centering\small
\caption{Carrier subsets in the joint matching (unseen objects, DA-V2, tuned settings). Trusted within = Within among the trusted pixels.}
\label{t:subset}
\setlength{\tabcolsep}{2pt}
\begin{tabular}{@{}lrrrr@{}}\toprule
Carriers & Within & MAE & Trusted & Trusted within\\\midrule
\hl $x_{10}{+}x_{16}{+}y_{10}{+}y_{16}$ (used) & 86.7\% & 2.47 & 91.2\% & 92.2\%\\
$x_{16}{+}y_{10}{+}y_{16}$ & 76.6\% & 4.44 & 92.9\% & 79.9\%\\
$x_{10}{+}x_{16}{+}y_{16}$ & 68.0\% & 5.87 & 89.3\% & 72.9\%\\
$x_{16}{+}y_{16}$ & 58.9\% & 7.24 & 90.2\% & 62.5\%\\
$x_{10}{+}y_{10}$ & 42.6\% & 8.05 & 83.6\% & 48.2\%\\\bottomrule
\end{tabular}
\end{table}

\subsection{Carrier design}\label{s:carrier}
\noindent Five carrier sets are compared on synthetic data. Each set has its own rendering of the training scenes (150 images per scene source, about 900 images), its own number of phase channels, and 8 epochs of training from the crossed-fringe model. The matcher (prior 15\,mm, phase noise 0.6\,rad) runs on the 72 synthetic test images, illuminated pixels. The training is short, so only the order of the designs is meaningful, not the absolute values (\cref{t:carrier}).
\begin{itemize}
\item \textbf{Longer periods are better.} 12/20 projector pixels gives the lowest MAE90 (2.10\,mm) and the fewest wrong-order pixels, 10.67/16 is in the middle and 8/12 is clearly worse (MAE90 5.79\,mm).
\item \textbf{The $y$ carriers add information.} Using only the $x$ carriers is the worst (MAE90 9.45\,mm), although the baseline lies mainly along $x$.
\item \textbf{Three carriers} lie between the reference and the best four-carrier design in the share within 1\,mm, but have the higher MAE.
\end{itemize}
The pattern in the experiments uses 10.67/16. A longer pair was not tested on real hardware.

\begin{table}[!t]
\centering\small
\caption{Carrier design (synthetic test images, illuminated pixels). Periods in projector pixels; wrong = error above 5\,mm.}
\label{t:carrier}
\setlength{\tabcolsep}{2pt}
\begin{tabular}{@{}lrrrrr@{}}\toprule
Carriers & Coarse MAE & Within & MAE & MAE90 & Wrong\\\midrule
$x,y$ at 12/20 & 22.24 & 68.3\% & 8.97 & 2.10 & 19.9\%\\
$x_{10.67}{+}x_{16}{+}y_{16}$ & 23.31 & 66.2\% & 12.36 & 3.85 & 23.6\%\\
\hl $x,y$ at 10.67/16 (used) & 23.69 & 59.4\% & 11.15 & 4.18 & 27.6\%\\
$x,y$ at 8/12 & 22.75 & 51.7\% & 13.15 & 5.79 & 36.6\%\\
$x$ at 10.67/16 only & 21.90 & 50.6\% & 16.82 & 9.45 & 42.3\%\\\bottomrule
\end{tabular}
\end{table}

\section{Consistency, trust and fill}\label{s:trust}
\gist{The trust test needs only a threshold in millimetres and is insensitive to it. The edge dead zone of the first pipeline version is switched off because it blocked corrections where no depth step exists.}

\subsection{Trust rule}
\noindent \Cref{t:trust} varies the rule that defines the trusted mask (dilated variant; for the final network the rule ``within 1\,mm of the median of the $9{\times}9$ neighbourhood'' gives 91.2\% trusted pixels on the unseen objects, 92.2\% of them within 1\,mm and 7.4\% above 5\,mm). The neighbour rule is insensitive to its threshold: from 0.25 to 4\,mm the trusted share goes from 71.7\% to 92.6\% and the share within 1\,mm of the trusted pixels from 94.5\% to 92.2\%. Ranking the pixels by the matcher margin (gap to the second-best minimum) gives a steeper trade-off: the best 9.4\% of the pixels are 98.6\% within 1\,mm. The margin rule needs a share chosen per scan, while the neighbour rule needs only a threshold in millimetres.

\begin{table}[!t]
\centering\small
\caption{Trust rules on unseen objects (dilated variant, tuned settings). Trusted share, within 1\,mm and MAE90 (mm) of the trusted pixels, and trusted pixels above 5\,mm.}
\label{t:trust}
\setlength{\tabcolsep}{2pt}
\begin{tabular}{@{}lrrrr@{}}\toprule
Rule & Trusted & Within & MAE90 & $>$5\,mm\\\midrule
within 0.25\,mm of the median & 71.7\% & 94.5\% & 0.04 & 5.4\%\\
\hl within 1\,mm of the median (used) & 89.3\% & 93.2\% & 0.05 & 6.6\%\\
within 4\,mm of the median & 92.6\% & 92.2\% & 0.05 & 7.5\%\\
margin, best 47\% & 47.0\% & 97.8\% & 0.03 & 2.2\%\\
margin, best 28\% & 28.2\% & 98.3\% & 0.03 & 1.7\%\\
margin, best 9.4\% & 9.4\% & 98.6\% & 0.03 & 1.4\%\\\bottomrule
\end{tabular}
\end{table}

\subsection{Clean-up and the edge dead zone}
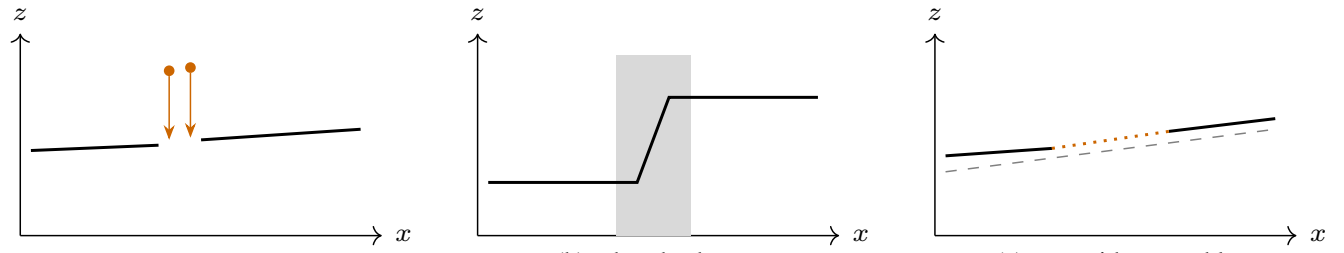
\begin{figure}[!t]
\centering
\resizebox{\columnwidth}{!}{%
\begin{tikzpicture}[font=\scriptsize,arr/.style={-{Stealth[length=1.4mm]}}]
% (a) order consistency
\begin{scope}
\draw[->] (0,0)--(3.4,0) node[right] {$x$}; \draw[->] (0,0)--(0,1.9) node[above] {$z$};
\draw[thick] (0.1,0.8)--(1.3,0.85) (1.7,0.9)--(3.2,1.0);
\foreach \p in {(1.4,1.55),(1.6,1.58)}{\fill[orange!80!black] \p circle (0.05);}
\draw[arr,orange!80!black] (1.4,1.5)--(1.4,0.9); \draw[arr,orange!80!black] (1.6,1.53)--(1.6,0.92);
\node[align=center] at (1.7,-0.55) {(a) wrong order,\\66\,mm off the neighbours,\\re-matched near their median};
\end{scope}
% (b) dead zone
\begin{scope}[xshift=4.3cm]
\draw[->] (0,0)--(3.4,0) node[right] {$x$}; \draw[->] (0,0)--(0,1.9) node[above] {$z$};
\fill[black!15] (1.3,0)rectangle(2.0,1.7);
\draw[thick] (0.1,0.5)--(1.5,0.5)--(1.8,1.3)--(3.2,1.3);
\node[align=center] at (1.7,-0.55) {(b) edge dead zone\\(first version): 95\,\% of it\\had no step; switched off};
\end{scope}
% (c) fill
\begin{scope}[xshift=8.6cm]
\draw[->] (0,0)--(3.4,0) node[right] {$x$}; \draw[->] (0,0)--(0,1.9) node[above] {$z$};
\draw[gray,dashed] (0.1,0.6)--(3.2,1.0);
\draw[thick] (0.1,0.75)--(1.1,0.82) (2.2,0.98)--(3.2,1.1);
\draw[orange!80!black,thick,dotted] (1.1,0.82)--(2.2,0.98);
\node[align=center] at (1.7,-0.55) {(c) gap without usable\\phase: coarse depth (dashed)\\plus a smooth correction (dotted)};
\end{scope}
\end{tikzpicture}}
\caption{Clean-up stages after the matching (schematic depth profiles along a row; thick: matched depth).}
\label{f:clean}
\end{figure}

\noindent \Cref{f:clean} sketches the three clean-up cases. Wrong-order pixels are re-matched near the median of their neighbours; the gaps without usable phase are filled from the coarse depth plus a smooth correction.

\noindent\textbf{Edge dead zone.} The first version of the pipeline blocked re-matching and fill inside a zone around the edges of the network coarse depth. It covers 16.2\% of the illuminated pixels, 94.3\% of which have no depth step in the reference, and it misses 46\% of the real steps (dilated variant, default matcher, 41 scans). The error inside the zone is 12.4\,mm with the dead zone and 7.1\,mm without; outside the zone it is unchanged (3.92 and 3.91\,mm). Raising the edge threshold shrinks the zone but never improves on switching it off.

\noindent\textbf{Un-illuminated region.} The dead zone also limits the fill across edges in the un-illuminated region, which the real scans cannot score. \Cref{t:dz} compares the fill with the dead zone on and off on the 72 synthetic test images, where the ground truth covers the un-illuminated pixels (here the pixels that are not within 4 camera pixels of the illuminated region). For the DA-V2 the dead zone helps in the dark and hurts in the light. With it, 17.7\% of the un-illuminated pixels are within 1\,mm against 7.4\% without it, but on the illuminated pixels the result is 70.9\% against 73.1\%. The tuned matcher uses no dead zone because the tuning objects are scored on the illuminated pixels. Sub-millimetre accuracy is not reached in the dark in either case (MAE 21.7 and 23.6\,mm). The synthetic-stage U-Net (the real-stage U-Net does not work on synthetic images) has a much worse coarse depth before the real stage, and the fill cannot repair this: 0.5\% within 1\,mm in the un-illuminated region. The ranking on the illuminated pixels is the same (off 53.8\%, on 50.0\%). The U-Net result says how much the real-data stage contributes, not how the final U-Net behaves in the dark.

\begin{table}[!t]
\centering\small
\caption{Edge dead zone of the fill on synthetic test images (tuned matcher, synthetic-stage models). Within = share within 1\,mm; MAE in mm.}
\label{t:dz}
\setlength{\tabcolsep}{2pt}
\begin{tabular}{@{}llrrrr@{}}\toprule
Network & Fill & \multicolumn{2}{c}{Illuminated} & \multicolumn{2}{c}{Un-illuminated}\\
 & & Within & MAE & Within & MAE\\\midrule
DA-V2 & none (coarse) & 4.1\% & 18.57 & 16.0\% & 24.39\\
 & dead zone on & 70.9\% & 8.96 & 17.7\% & 21.72\\
 & dead zone off & 73.1\% & 7.89 & 7.4\% & 23.57\\\midrule
U-Net & none (coarse) & 2.0\% & 53.30 & 0.5\% & 204.20\\
 & dead zone on & 50.0\% & 45.88 & 0.5\% & 204.19\\
 & dead zone off & 53.8\% & 41.91 & 0.6\% & 202.93\\\bottomrule
\end{tabular}
\end{table}

\section{Seeds, networks and held-out objects}\label{s:seeds}
\gist{Seed variation is far inside the interval over scans. A U-Net without pre-training matches the DA-V2 on typical pixels of the unseen objects, but both networks fail on a new object when the training set contains only three others.}

\subsection{Seeds and training length}
\noindent \Cref{t:seed} repeats the real-data stage with two more seeds for DA-V2 and for the dilated variant, and once with 300 epochs for the dilated variant. The three DA-V2 seeds agree within 0.3\% on the unseen objects (86.7--87.0\%) and on all 41 scans (91.6--91.7\%), far inside the bootstrap interval of $\pm6$\%. The dilated variant behaves alike (85.9--86.2\% on the unseen objects). 300 epochs of the dilated variant gain 1.8\% on all scans, which include the training objects, and nothing on the unseen objects (86.0\%).

\begin{table}[!t]
\centering\small
\caption{Seeds and training length (tuned matcher, 2000 bootstrap draws over scans; DA-V2 = U-Net phase CNN). Unseen and All = share within 1\,mm on the unseen objects and on all scans, each followed by its MAE (mm).}
\label{t:seed}
\setlength{\tabcolsep}{2pt}
\begin{tabular}{@{}lrrrrr@{}}\toprule
Run & Unseen & [95\%] & MAE & All & MAE\\\midrule
\hl DA-V2, seed 0 (main) & 86.7\% & [79.8, 92.5] & 2.47 & 91.7\% & 1.15\\
DA-V2, seed 1 & 86.7\% & [79.9, 92.3] & 2.46 & 91.6\% & 1.14\\
DA-V2, seed 2 & 87.0\% & [80.4, 92.6] & 2.40 & 91.7\% & 1.14\\
\midrule
Dilated, seed 0, 200 ep & 85.9\% & [79.0, 91.5] & 2.53 & 89.7\% & 1.30\\
Dilated, seed 1, 200 ep & 86.2\% & [79.3, 91.8] & 2.38 & 89.7\% & 1.28\\
Dilated, seed 2, 200 ep & 86.1\% & [79.2, 91.9] & 2.42 & 89.8\% & 1.28\\
Dilated, seed 0, 300 ep & 86.0\% & [79.3, 91.5] & 2.55 & 91.5\% & 1.16\\\bottomrule
\end{tabular}
\end{table}

\subsection{Networks without pre-training}\label{s:arch}
\noindent Two networks without pre-training replace the DA-V2. They have the same outputs (depth, four wrapped phases, projector coordinates), the same data and the same real stage of 200 epochs, and use the tuned matcher. The U-Net (69.8M parameters) trains 16 synthetic epochs, twice the DA-V2 budget, because it has no pre-trained weights and no crossed-fringe stage. The ViT (40.6M parameters, 12 layers of width 512, patch 32 at stride 16, a convolutional head that also sees the input image) uses the same schedule.

\begin{table}[!t]
\centering\small
\caption{Network comparison (tuned matcher, 95\% intervals over scans). Both DA-V2 rows are pre-trained. Unseen and All = share within 1\,mm on the unseen objects and on all 41 scans, each followed by its MAE (mm); Trusted = trusted share on the unseen objects.}
\label{t:arch}
\setlength{\tabcolsep}{2pt}\footnotesize
\begin{tabular}{@{}lrrrrrr@{}}\toprule
Network & Unseen & [95\%] & MAE & Trusted & All & MAE\\\midrule
\hl DA-V2, U-Net phase CNN & 86.7\% & [79.8, 92.5] & 2.47 & 91.2\% & 91.7\% & 1.15\\
DA-V2, dilated phase CNN & 85.9\% & [79.0, 91.5] & 2.53 & 89.3\% & 89.7\% & 1.30\\
U-Net (scratch) & 89.5\% & [83.9, 93.9] & 2.82 & 93.1\% & 94.2\% & 1.16\\
ViT (scratch) & 9.3\% & [3.6, 15.2] & 25.77 & 15.1\% & 18.4\% & 18.19\\\bottomrule
\end{tabular}
\end{table}

\noindent\textbf{The U-Net matches the DA-V2.} The U-Net reaches 89.5\% within 1\,mm on the unseen objects, against 86.7\% for the DA-V2. The intervals overlap, so the nine unseen scans do not resolve this difference. Its coarse depth is no better (MAE 10.4\,mm against 9.3\,mm) and its MAE is not lower (2.82 against 2.47\,mm). The sub-millimetre share therefore comes from the phases and the matcher, not from the pre-trained backbone. \Cref{t:unetseed} gives three seeds. The share within 1\,mm varies by 0.7\%, but the MAE varies from 2.8 to 5.1\,mm on the unseen objects (DA-V2 seeds: 2.4 to 2.5\,mm), because a few wrong-order pixels dominate it.

\begin{table}[!t]
\centering\small
\caption{U-Net seeds (tuned matcher, 95\% intervals over scans). Unseen and All = share within 1\,mm, followed by the MAE (mm).}
\label{t:unetseed}
\setlength{\tabcolsep}{2pt}
\begin{tabular}{@{}lrrrrr@{}}\toprule
Run & Unseen & [95\%] & MAE & All & MAE\\\midrule
Seed 0 & 89.5\% & [83.9, 93.9] & 2.82 & 94.2\% & 1.16\\
Seed 1 & 89.8\% & [84.2, 93.9] & 5.13 & 94.2\% & 1.73\\
Seed 2 & 89.1\% & [83.3, 93.5] & 3.73 & 94.0\% & 1.39\\\bottomrule
\end{tabular}
\end{table}

\noindent\textbf{The ViT underfits.} It reaches 9.3\% within 1\,mm on the unseen objects and 18.4\% on all scans. It also fails on the objects it trained on (8.6\% for the angel, 17.5\% for the bell), and its coarse depth is poor (MAE 28.4\,mm on the unseen objects). Its real-stage training loss stays at 7.4, against 0.45 for the U-Net at the same stage. The schedule (16 synthetic and 200 real epochs, U-Net learning rates) was not tuned for a transformer without pre-training. The result shows that the pipeline needs a network that fits the phases, and that a from-scratch transformer with this schedule does not. It is not evidence about transformers in general.

\subsection{Leave-one-object-out}\label{s:looo}
\noindent \Cref{t:looo} removes one tuning object from the real training set (dilated variant) and tests it on its 8 scans, with the matcher settings unchanged. The angel, restoration piece and bell fall from 80--91\% to 49--55\% within 1\,mm, and their MAE rises from 1--2\,mm to 9--13\,mm. The figurine falls only from 94.7\% to 89.1\%. The unseen fossil and benchmark change little (82.6--85.5\% against 85.9\%), so no single tuning object carries the unseen result. The held-out tuning objects are much worse than the unseen ones; the experiment does not separate the difficulty of these objects from the missing training scans. The figurine run has no validation scans left, because its validation positions are the evaluation scans. It uses the last checkpoint instead of the best one, and the other three runs select the best checkpoint as usual.

\begin{table}[!t]
\centering\small
\caption{Leave-one-object-out (dilated variant, tuned matcher). Each row is a model trained without the named object. Within and MAE (mm) are those of the left-out object; Main is the share within 1\,mm of the dilated model that trained on it (the figurine on its other positions); Unseen and its MAE are for fossil and benchmark.}
\label{t:looo}
\setlength{\tabcolsep}{2pt}
\begin{tabular}{@{}lrrrrr@{}}\toprule
Left out & Within & MAE & Main & Unseen & MAE\\\midrule
Angel & 49.3\% & 8.51 & 80.5\% & 83.8\% & 3.51\\
Restoration & 54.3\% & 13.13 & 90.8\% & 85.5\% & 2.53\\
Bell & 54.8\% & 9.14 & 91.4\% & 84.8\% & 2.78\\
Figurine & 89.1\% & 2.18 & 94.7\% & 82.6\% & 3.42\\\bottomrule
\end{tabular}
\end{table}

\noindent\textbf{Scratch U-Net.} \Cref{t:unetlooo} repeats the test for the U-Net. On the objects in the training set the U-Net is more accurate than the DA-V2: it reaches 93.9\%, 95.6\%, 95.2\% and 96.7\% within 1\,mm for angel, restoration piece, bell and figurine, against 85.5\%, 93.8\%, 92.7\% and 95.8\% for the DA-V2 (dilated: 80.5\%, 90.8\%, 91.4\% and 94.7\%). When the object is removed, both networks fall. The U-Net falls further on three of four objects (the bell to 30\% with an MAE of 27.8\,mm) and is better on one (restoration piece, 69\% against 54\%). Averaged over the four objects, the left-out share is 52\% for the U-Net and 62\% for the dilated DA-V2. With the figurine removed, the U-Net also loses accuracy on the unseen objects (54.7\%, interval 27.6--81.6\%); that run keeps its last checkpoint. The pre-trained model generalises somewhat more reliably, but neither network generalises to a new object. The limit is the training set, not one architecture.

\begin{table}[!t]
\centering\small
\caption{Leave-one-object-out, U-Net against DA-V2 (dilated variant; tuned matcher). Left out = within-1\,mm share and MAE (mm) of the removed object; Unseen = fossil and benchmark.}
\label{t:unetlooo}
\setlength{\tabcolsep}{2pt}
\begin{tabular}{@{}llrrrr@{}}\toprule
Left out & Network & Left out & MAE & Unseen within & Unseen MAE\\\midrule
Angel & U-Net & 42.3\% & 24.85 & 83.4\% & 8.86\\
 & DA-V2 & 49.3\% & 8.51 & 83.8\% & 3.51\\
Restoration & U-Net & 69.1\% & 9.69 & 87.0\% & 6.31\\
 & DA-V2 & 54.3\% & 13.13 & 85.5\% & 2.53\\
Bell & U-Net & 30.2\% & 27.81 & 85.5\% & 3.91\\
 & DA-V2 & 54.8\% & 9.14 & 84.8\% & 2.78\\
Figurine & U-Net & 67.4\% & 10.09 & 54.7\% & 20.12\\
 & DA-V2 & 89.1\% & 2.18 & 82.6\% & 3.42\\\bottomrule
\end{tabular}
\end{table}

\section{Network phases and where they fail}\label{s:phases}
\gist{Classical Fourier-transform phases lose about 30\%, the wrapped phase is wrong over the whole image rather than at particular pixels, and ideal phases would remove four fifths of the wrong-order pixels.}

\subsection{Classical phases instead of network phases}\label{s:ftp}
\noindent The network phases are replaced by classical Fourier-transform profilometry (FTP) on the same image. The rectified single-shot image is demodulated at the four carrier frequencies, which are the median gradient of the projector coordinate over the first six angel scans divided by the period, after removing a 20\,px Gaussian background. The low-pass width is 4\,px, chosen from a sweep of 12, 7, 4 and 2.5\,px (32, 44, 50 and 39\% on three angel scans). The offset of every carrier is fitted on the same angel scans as for the network. The coarse depth, the matcher, the consistency step and the fill are those of the method. \Cref{t:ftp} shows that the FTP phases give far fewer pixels within 1\,mm. The median phase error on the angel scans is 0.45--0.90\,rad for FTP against 0.13--0.29\,rad for the network. With 4.7 to 7 camera pixels per carrier period, a linear filter cannot separate the carriers well, which is the case the network phases are trained for.

\begin{table}[!t]
\centering\small
\caption{Phase source of the matcher (DA-V2, tuned matcher, same coarse depth). Wrong = final error above 5\,mm.}
\label{t:ftp}
\setlength{\tabcolsep}{2pt}
\begin{tabular}{@{}llrrrr@{}}\toprule
Scans & Phase source & Within & MAE & MAE90 & Wrong\\\midrule
\hl Unseen & Network & 86.7\% & 2.47 & 0.26 & 11.9\%\\
 & FTP & 58.2\% & 5.19 & 2.81 & 28.1\%\\
 & FTP, no consistency or fill & 51.6\% & 7.22 & 4.70 & 39.4\%\\\midrule
\hl All (41) & Network & 91.7\% & 1.15 & 0.06 & 6.4\%\\
 & FTP & 57.0\% & 3.41 & 1.81 & 22.6\%\\
 & FTP, no consistency or fill & 49.0\% & 5.36 & 3.37 & 34.8\%\\\bottomrule
\end{tabular}
\end{table}

\subsection{Training changes for the wrapped phase}
\noindent Two changes target the wrapped phase, the main source of wrong-order pixels, with a real stage from the same synthetic model as the main run (dilated variant). The first weights the phase loss by the local fringe contrast (local standard deviation of the input, relative to its median, at most 1). The second adds a residual refinement stage after the phase head (three convolutions of width 32, the last one initialised to zero). \Cref{t:phq} shows that neither changes the result. The shares within 1\,mm differ by at most 0.6\% on all scans and 0.2\% on the unseen objects, inside the intervals. The phase error does not move either: the pixels with the largest error (depth edges, dark pixels) stay at the same level. The limit on the wrapped phase is therefore not the loss weighting and not the capacity of the head.

\begin{table}[!t]
\centering\small
\caption{Changes to the phase training (dilated variant, tuned matcher). Unseen and All = share within 1\,mm on the unseen objects and on all scans, each followed by its MAE (mm). Edge, dark = share of the depth-edge and the dark pixels with a phase error above 0.5\,rad on at least one carrier (all scans).}
\label{t:phq}
\setlength{\tabcolsep}{2pt}\footnotesize
\begin{tabular}{@{}lrrrrrr@{}}\toprule
Run & Unseen & MAE & All & MAE & Wrong & Edge, dark\\\midrule
\hl Main (dilated) & 85.9\% & 2.53 & 89.7\% & 1.30 & 7.6\% & 82\%, 33\%\\
Contrast weight & 85.7\% & 2.49 & 89.8\% & 1.27 & 7.4\% & 82\%, 34\%\\
Refinement stage & 85.9\% & 2.50 & 90.3\% & 1.25 & 7.3\% & 81\%, 33\%\\\bottomrule
\end{tabular}
\end{table}

\subsection{Where the wrapped phase fails}\label{s:phase}
\noindent The phase error is the wrapped difference between the network phase (offset-corrected) and the ideal phase computed from the reference and the calibration. \Cref{t:phase} gives it per pixel group (DA-V2, default matcher; groups overlap).
\begin{itemize}
\item \textbf{The error is spread over the whole image.} The median is 0.12--0.21\,rad and one pixel in four is above 0.5\,rad in at least one carrier.
\item \textbf{Few wrong-order pixels sit at special places.} Low-modulation pixels (phase magnitude below 0.25 in any carrier, not matched) hold about 15\% of all wrong-order pixels and depth edges about 4\%. About 81\% lie in normal pixels with a moderate phase error.
\item \textbf{Surface slope is not a cause.} Steep pixels ($>50^\circ$, 27.5\% of the pixels) have the same phase error as all pixels.
\item \textbf{Dark and bright pixels are bad on some objects only.} Dark pixels on the benchmark (0.43\,rad, 32\% wrong order), bright pixels on the fossil (0.34\,rad, 28\%) and the bell (0.33\,rad, 39\%). The cause is probably material and shine.
\item \textbf{The 16-pixel carriers are better} than the 10.67-pixel ones, as expected for a finer pattern.
\end{itemize}

\begin{table}[!t]
\centering\small
\caption{Median phase error (rad) per carrier and wrong-order share, per pixel group (DA-V2, default matcher, 41 scans).}
\label{t:phase}
\setlength{\tabcolsep}{2pt}
\begin{tabular}{@{}lrrrrrr@{}}\toprule
Group & Share & $x_{10}$ & $x_{16}$ & $y_{10}$ & $y_{16}$ & Wrong\\\midrule
All illuminated & 100\% & 0.18 & 0.12 & 0.21 & 0.13 & 15.9\%\\
Low modulation & 3.9\% & 1.17 & 0.47 & 0.61 & 0.43 & 60.3\%\\
Depth edge & 1.7\% & 0.51 & 0.35 & 0.48 & 0.35 & 37.4\%\\
Steep ($>50^\circ$) & 27.5\% & 0.16 & 0.11 & 0.21 & 0.13 & 15.0\%\\
Flat ($<20^\circ$) & 1.4\% & 0.18 & 0.13 & 0.17 & 0.11 & 9.5\%\\
Darkest 15\% & 15.0\% & 0.19 & 0.13 & 0.21 & 0.13 & 15.4\%\\
Brightest 10\% & 10.0\% & 0.22 & 0.15 & 0.25 & 0.16 & 22.5\%\\\bottomrule
\end{tabular}
\end{table}

\subsection{Oracle swaps}
\noindent One input of the matcher is replaced with the reference (\cref{t:oracle}). Ideal phases remove about 80\% of the wrong-order pixels (15.9\% to 2.9\%). A reference coarse depth with 5\,mm noise lowers the share by only 1.5\%, and with 15\,mm noise it rises to 18.0\%. With both ideal phases and the reference coarse depth, 1.8\% remain. The phases, not the coarse depth, are the main limit.

\begin{table}[!t]
\centering\small
\caption{Oracle swaps (DA-V2, all 41 scans, default matcher). Wrong order: final error above 5\,mm.}
\label{t:oracle}
\setlength{\tabcolsep}{2pt}
\begin{tabular}{@{}lrrr@{}}\toprule
Matcher input & Wrong order & Within & MAE\\\midrule
\hl Network phases, network coarse depth & 15.9\% & 82.7\% & 5.17\\
Ideal phases & 2.9\% & 95.5\% & 0.62\\
Reference coarse depth (5\,mm noise) & 14.4\% & 83.8\% & 4.54\\
Reference coarse depth (15\,mm noise) & 18.0\% & 81.0\% & 6.09\\
Both (5\,mm noise) & 1.8\% & 96.1\% & 0.34\\\bottomrule
\end{tabular}
\end{table}

\section{Robustness and validation}\label{s:rob}
\gist{The calibration must be accurate to about one projector pixel, and the pipeline fails when the coarse depth is off by more than the alias spacing, as on a new rig. The network is sensitive to input noise in the range of its synthetic training, which noise-aware retraining largely removes.}

\subsection{Calibration sensitivity}\label{s:calib}
\noindent The reference is computed with the correct rig and the matching with a perturbed one (11 scans, dilated variant, tuned matcher; \cref{t:calib}). Unperturbed, 87.6\% of the pixels are within 1\,mm. The trusted share barely changes (79--88\%) while the accuracy falls, so the trust mask does not detect a wrong calibration.
\begin{itemize}
\item \textbf{Projector principal point.} Tolerant to 1--2\,px ($u$: 87.1\% and 81.5\% at $\pm2$\,px), and lost at 4\,px (40\% for $u$, 9--13\% for $v$).
\item \textbf{Projector focal length.} An error of 0.1\% costs at most 1.1\%, 0.3\% costs 15--19\%.
\item \textbf{Baseline length and camera principal point.} The most sensitive: a baseline error of 0.3\% leaves 19--25\%, and 1\,px at the camera leaves 32\%.
\item \textbf{Projector rotation.} 0.05$^\circ$ about $x$ or $y$ costs 11--22\%, about $z$ it costs 1 point.
\item \textbf{Projector distortion.} Ignoring it leaves 37.3\%.
\end{itemize}
The calibration must therefore be correct to about one projector pixel at the principal point and to better than 0.3\% in baseline and focal length.

\begin{table}[!t]
\centering\small
\caption{Calibration sensitivity: share within 1\,mm (\%) for perturbations of the matching rig (11 scans, dilated variant, tuned matcher; unperturbed 87.6\%). Pairs are negative/positive perturbation; rotations are positive only.}
\label{t:calib}
\footnotesize
\begin{tabular}{@{}p{0.36\columnwidth}p{0.59\columnwidth}@{}}\toprule
Perturbation & Amount: within 1\,mm\\\midrule
Projector principal point $u$ (px) & $\pm1$: 88.3/85.7; $\pm2$: 87.1/81.5; $\pm4$: 40.5/42.1; $\pm8$: 2.8/4.8\\
Projector principal point $v$ (px) & $\pm1$: 88.0/84.9; $\pm2$: 82.3/76.1; $\pm4$: 13.1/8.8; $\pm8$: 0.5/0.9\\
Projector focal length (relative) & $\pm0.1\%$: 86.5/87.3; $\pm0.3\%$: 72.2/68.7; $\pm1\%$: 14.1/12.5\\
Baseline length (relative) & $\pm0.3\%$: 18.8/24.8; $\pm1\%$: 0.7/0.8; $\pm3\%$: 1.2/0.7\\
Camera principal point (px) & $\pm1$: 32.3/31.7; $\pm2$: 0.8/1.0; $\pm4$: 0.4/0.4\\
Projector rotation about $x$ ($^\circ$) & 0.05: 65.2; 0.1: 1.0; 0.25: 27.6; 0.5: 1.1; 1: 5.2\\
Projector rotation about $y$ ($^\circ$) & 0.05: 76.5; 0.1: 6.6; 0.25: 7.0; 0.5: 3.1; 1: 7.1\\
Projector rotation about $z$ ($^\circ$) & 0.05: 86.5; 0.1: 84.3; 0.25: 61.5; 0.5: 23.4; 1: 7.1\\
Projector distortion ignored & 37.3\\\bottomrule
\end{tabular}
\end{table}

\subsection{A new rig: the vase}\label{s:vase}
\noindent The vase was scanned with a changed rig geometry (8 scans, median depth 0.44\,m against 0.35\,m for the training objects). \Cref{t:vase} shows that all three networks fail. The coarse depth is 73--82\,mm too near, far outside the $\pm33$\,mm that the four carriers tolerate (\cref{s:match}), and the network phases are worse as well (median error 0.52, 0.37, 0.52 and 0.35\,rad against 0.15, 0.11, 0.19 and 0.12\,rad on the training scans). The final result has at most 0.2\% of the pixels within 1\,mm, whether the matching uses the calibration of the vase or that of the training rig. The trusted share stays at 59--71\%, so the trust mask does not warn here. A different calibration alone does not cause the failure, which is the coarse depth; no architecture recovers where the coarse depth is off by more than the ambiguity period. The U-Net also shows the same sensitivity to the calibration as the DA-V2 (principal point error of 2 projector pixels: 0.3 to 7\% lost; 4 pixels: 47 to 50; 8 pixels: 2\% left), which follows from the matcher and not from the network.

\begin{table}[!t]
\centering\small
\caption{Scans of the vase on a new rig (tuned matcher, matching with the calibration of the vase). Coarse and final MAE in mm, Within = share within 1\,mm. Matching with the calibration of the training rig changes the final MAE by at most 1.2\,mm and Within by at most 0.1\%. Training rig: the 18 scans with the training configuration.}
\label{t:vase}
\setlength{\tabcolsep}{2pt}\footnotesize
\begin{tabular}{@{}lrrrrr@{}}\toprule
 & \multicolumn{3}{c}{Vase} & \multicolumn{2}{c}{Training rig}\\
Network & Coarse & Final & Within & Coarse & Within\\\midrule
DA-V2, U-Net phase CNN & 73.17 & 74.17 & 0.0\% & 5.95 & 91.1\%\\
DA-V2, dilated & 74.41 & 75.28 & 0.0\% & 6.67 & 89.5\%\\
U-Net (scratch) & 82.13 & 83.23 & 0.2\% & 5.82 & 93.1\%\\\bottomrule
\end{tabular}
\end{table}

\subsection{Input noise}\label{s:noise}
\noindent We test the response of the pipeline to input noise. Gaussian noise is added to the network input of all 41 scans and the pipeline is scored again (dilated variant, tuned matcher). The levels are multiples of 0.028 of the 0--1 range, the noise standard deviation of one exposure; the frame noise measured on the scans is 0.032 (10th to 90th percentile over carriers and scans: 0.009--0.070).
\begin{itemize}
\item \textbf{White noise (\cref{t:noise}).} A quarter of level 1 costs 2\% within 1\,mm on the unseen objects, half of it 11\%, and the full level 48\%. The trusted share falls with it, and the trusted pixels stay about as precise as before only while the trusted share holds.
\item \textbf{Noise of the synthetic stage (\cref{t:noisep}).} The synthetic stage already adds Poisson-Gaussian noise to 80\% of its images (150 to 900 photo-electrons at full scale, plus 0.3 to 1.5 grey levels). At the brightness of the real scans this is a standard deviation of about 0.01 to 0.026, of the order of level 1. The same noise model is added here on top of the real inputs, which carry their own noise, so the combined noise is larger than the added one. The accuracy falls inside this range too (unseen objects 79\% and 70\% at the two weakest settings, 40\% at 150) and collapses outside it.
\item \textbf{Cause.} The model that is trained last, on the real images, is therefore not robust to noise in the range of its own synthetic stage. A likely reason is that the real stage used only low-noise inputs; this was not tested directly.
\item \textbf{Noise-aware retraining (\cref{t:noiseaug}).} The real stage is retrained with the same Poisson-Gaussian noise on 80\% of its training inputs (gain 150 to 900, 0.3 to 1.5 grey levels; one run, seed 0, otherwise the same synthetic start and settings). On clean input the accuracy is about the same on the unseen objects (84.8\% against 85.9\%, MAE 2.90 against 2.53\,mm, inside the intervals) and 1.6\% lower on all scans (88.1\% against 89.7\%). At level 1 of white noise, a noise type not used in training, the unseen result is 76\% instead of 38\%, and at synthetic noise gains of 400 and 150 it is 83\% and 79\% instead of 70\% and 40\%. Beyond this range it still collapses (28\% at level 2). Noise-aware training therefore removes most of the loss at level 1, at a small cost on clean input.
\end{itemize}

\begin{table}[!t]
\centering\small
\caption{Extra Gaussian noise on the network input (dilated variant; standard deviation in the 0--1 range; level 1 = 0.028). Within = share within 1\,mm; MAE in mm; Trusted = trusted share.}
\label{t:noise}
\setlength{\tabcolsep}{2pt}
\begin{tabular}{@{}rrrrrrrr@{}}\toprule
 & & \multicolumn{3}{c}{Unseen objects} & \multicolumn{3}{c}{All 41 scans}\\
Level & Noise & Within & MAE & Trusted & Within & MAE & Trusted\\\midrule
0 & 0 & 85.9\% & 2.53 & 89.3\% & 89.7\% & 1.30 & 88.3\%\\
0.1 & 0.003 & 85.6\% & 2.58 & 89.2\% & 89.5\% & 1.33 & 88.2\%\\
0.25 & 0.007 & 83.9\% & 2.95 & 88.3\% & 88.3\% & 1.50 & 87.3\%\\
0.5 & 0.014 & 74.4\% & 5.10 & 82.0\% & 82.1\% & 2.53 & 82.7\%\\
1 & 0.028 & 38.3\% & 10.38 & 42.2\% & 60.1\% & 5.03 & 60.4\%\\
2 & 0.056 & 13.9\% & 20.65 & 27.9\% & 25.9\% & 13.11 & 34.3\%\\
4 & 0.111 & 8.0\% & 24.40 & 21.5\% & 9.0\% & 19.70 & 21.1\%\\\bottomrule
\end{tabular}
\end{table}

\begin{table}[!t]
\centering\small
\caption{Poisson-Gaussian noise of the synthetic augmentation added to the real network input (dilated variant; gain = photo-electrons at full scale; Gaussian part 0.9 grey levels). Within = share within 1\,mm; MAE in mm; Trusted = trusted share.}
\label{t:noisep}
\setlength{\tabcolsep}{2pt}
\begin{tabular}{@{}rrrrrrr@{}}\toprule
 & \multicolumn{3}{c}{Unseen objects} & \multicolumn{3}{c}{All 41 scans}\\
Gain & Within & MAE & Trusted & Within & MAE & Trusted\\\midrule
none & 85.9\% & 2.53 & 89.3\% & 89.7\% & 1.30 & 88.3\%\\
900 & 79.2\% & 3.95 & 84.9\% & 83.4\% & 2.05 & 83.0\%\\
400 & 69.6\% & 5.10 & 74.9\% & 74.5\% & 2.92 & 74.5\%\\
150 & 40.2\% & 8.33 & 41.4\% & 51.2\% & 5.45 & 51.5\%\\
75 & 24.7\% & 11.29 & 28.2\% & 32.7\% & 8.27 & 35.9\%\\\bottomrule
\end{tabular}
\end{table}

\begin{table}[!t]
\centering\small
\caption{Real stage retrained with Poisson-Gaussian noise on its training inputs (dilated variant, one seed), tested with the noise of \cref{t:noise,t:noisep}. White: level in multiples of 0.028; gain: photo-electrons at full scale. Within = share within 1\,mm; MAE in mm; Trusted = trusted share.}
\label{t:noiseaug}
\setlength{\tabcolsep}{2pt}
\begin{tabular}{@{}lrrrrrr@{}}\toprule
 & \multicolumn{3}{c}{Unseen objects} & \multicolumn{3}{c}{All 41 scans}\\
Test noise & Within & MAE & Trusted & Within & MAE & Trusted\\\midrule
clean & 84.8\% & 2.90 & 88.8\% & 88.1\% & 1.51 & 86.8\%\\
white 0.25 & 84.1\% & 3.05 & 88.3\% & 87.5\% & 1.58 & 86.3\%\\
white 0.5 & 82.7\% & 3.34 & 87.0\% & 86.2\% & 1.73 & 84.9\%\\
white 1 & 76.2\% & 4.06 & 79.0\% & 80.6\% & 2.23 & 78.8\%\\
white 2 & 27.9\% & 10.26 & 28.3\% & 49.5\% & 5.77 & 47.2\%\\
white 4 & 13.4\% & 17.46 & 20.5\% & 16.4\% & 12.81 & 20.2\%\\
gain 900 & 83.8\% & 3.08 & 87.9\% & 86.7\% & 1.64 & 85.4\%\\
gain 400 & 83.1\% & 3.10 & 86.8\% & 85.2\% & 1.75 & 83.7\%\\
gain 150 & 79.0\% & 3.45 & 81.5\% & 80.0\% & 2.15 & 78.0\%\\
gain 75 & 65.1\% & 4.38 & 64.8\% & 69.0\% & 2.94 & 65.9\%\\\bottomrule
\end{tabular}
\end{table}

\subsection{Independent check against CAD}\label{s:cad}
\noindent The reference shares its phase source with the method. The benchmark (one position) is therefore compared with the CAD model of its orbs and plate after rigid alignment (68,070 reference pixels with a CAD hit within 10\,mm, 68\% of all reference pixels).
\begin{itemize}
\item \textbf{Reference against CAD.} On the orbs the MAE is 0.20\,mm (median 0.16\,mm, 99.4\% within 1\,mm). On the plate top the reference and the CAD differ by a median of 2.1\,mm (27\% within 1\,mm). It is not known whether the CAD plate height or the reference is wrong.
\item \textbf{Method against CAD (tuned).} On the orbs, 55\% of the pixels are within 1\,mm and the median error is 0.39\,mm; the rest are mostly wrong-order pixels (90th percentile 58\,mm). On the plate top, 4\% are within 1\,mm and the median error is 13.7\,mm, because most of the plate has a wrong fringe order (coarse error about 20\,mm there). Over all CAD pixels the median error is 7.3\,mm, and 0.33\,mm on the trusted pixels (58\% within 1\,mm).
\end{itemize}
Where the fringe order is right, the method is within about a millimetre of the CAD (median 0.39\,mm on the orbs against 0.16\,mm for the reference). The plate, a large flat surface, shows the failure mode of a wrong fringe order.

\subsection{Un-illuminated region and run time}
\noindent\textbf{Un-illuminated region.} On synthetic scans the field correction lowers the un-illuminated MAE by 7\% (22.97 to 21.43\,mm); a scale correction makes it worse. Scene 1 is better (12.46 to 8.93\,mm).

\noindent\textbf{Run time.} The network takes 0.15\,s, and matching, consistency and fill take 0.27\,s per image (median over 6 scans with a median of 38k matched pixels per image; input $832{\times}1024$; RTX 4080). The network needs 1.55\,GB of memory.

\section{Training settings}\label{s:settings-train}
\noindent Optimiser AdamW; cosine schedule with one warm-up epoch; gradient clipping 1.0; batch size 2; EMA decay 0.999; input $832{\times}1024$. Synthetic stage 2: head learning rate $5{\times}10^{-5}$, backbone one tenth of that. Real stage: $2{\times}10^{-5}$, backbone factor 0.1, 200 epochs, masked L1 on pixels with a reference, phase weight 2.0, order weight 0.5, checkpoint selected on the illuminated MAE of the validation scans. The phase offset of every carrier is fitted on 6 scans of the angel, the first tuning object.

\noindent\textbf{Release.} The synthetic data generator, the trained models and the evaluation code will be released upon acceptance.

{\small

}

\end{document}